\documentclass[letterpaper, 10 pt, conference]{ieeeconf}  

\IEEEoverridecommandlockouts                              

\makeatletter
\newcommand*{\rom}[1]{\expandafter\@slowromancap\romannumeral #1@}
\makeatother

\usepackage{graphics} 
\usepackage{graphicx}
\usepackage{epsfig} 
\usepackage{epstopdf} 
\usepackage{amsmath} 
\usepackage{amssymb}  

\usepackage{microtype}
\usepackage[usenames,dvipsnames,table]{xcolor}
\usepackage{commath}
\usepackage{todonotes}
\usepackage{bm}
\usepackage{mathtools}

\usepackage[utf8]{inputenc}
\usepackage[english]{babel}
\PassOptionsToPackage{hyphens}{url}\usepackage[colorlinks, citecolor={blue}, linkcolor={blue}]{hyperref}
\usepackage{import}
\usepackage{paralist}
\usepackage{algpseudocode}
\usepackage{algorithm}
\usepackage{gensymb}
\usepackage{dsfont}
\usepackage{siunitx}
\usepackage[bottom]{footmisc}
\usepackage{framed}
\usepackage{balance} 
\usepackage{caption}
\usepackage[skins]{tcolorbox}  
\usepackage{subcaption}

\usepackage{todonotes}

\DeclareCaptionFont{mysize}{\fontsize{8}{10}\selectfont}
\usepackage{pifont}

\usepackage{colortbl}	
\usepackage{xspace}

\usepackage{mathrsfs}
\usepackage{siunitx}

\newtheorem{assumption}{Assumption}

\newtheorem{remark}{Remark}

\newtheorem{definition}{Definition}

\newcommand\I[1]{\mathbf{I}_{#1}}
\newcommand\Zero[1]{\mathbf{0}_{#1}}
\newcommand\Zeros[2]{\mathbf{0}_{#1 \times #2}}

\newcommand\X{\mathbf{X}}
\newcommand\Y{\mathbf{Y}}
\newcommand\Z{\mathbf{Z}}

\newcommand\A{\mathbf{A}}

\newcommand\aVec{\mathbf{a}}

\newcommand\uVec{\mathbf{u}}
\newcommand\vVec{\mathbf{v}}

\newcommand\R{\mathbb{R}}

\newcommand\ghat[2]{\hm{\left[} #2 \hm{\right]}^\wedge_{#1}}
\newcommand\gvee[2]{\hm{\left[} #2 \hm{\right]}^\vee_{#1}}
\newcommand\gexp[1]{\exp_{#1}}
\newcommand\glog[1]{\log_{#1}}
\newcommand\gexphat[1]{\exp_{#1}^\wedge}

\newcommand\glogvee[1]{\log_{#1}^\vee}
\newcommand\gadj[1]{\textup{Ad}_{#1}}

\newcommand\gljac[1]{\mathbf{J}^l_{#1}}
\newcommand\grjac[1]{\mathbf{J}^r_{#1}}

\newcommand\SO[1]{\textup{SO}(#1)}
\newcommand\so[1]{\mathfrak{so}(#1)}
\newcommand\SE[1]{\textup{SE}(#1)}

\newcommand\SEk[2]{\textup{SE}_{#1}(#2)}

\newcommand\Exp{\textup{Exp}}

\newcommand\bias{{\mathbf{b}}}

\newcommand\noiseBias[1]{{\mathbf{w}_{{\mathbf{b}}}^{#1}}}

\newcommand\acc[2]{{\prescript{{#2}}{}{\alpha}^{\text{lin}}_{#1, #2}}}
\newcommand\accBar[2]{{\prescript{{#2}}{}{\bar{\alpha}}^{\text{lin}}_{#1, #2}}}
\newcommand\yAcc[2]{ { { \tilde{\alpha} }^{g, \text{lin}}_{{#1, #2}}  } }
\newcommand\biasAcc{{{\mathbf{b}}_{a}}}
\newcommand\biasAccHat{{\mathbf{\hat{b}}_a}}
\newcommand\noiseAcc[1]{{\mathbf{w}_{a}^{#1}}}

\newcommand\yGyro[2]{{\prescript{{#2}}{}{\tilde{\omega}}_{#1, #2}}}
\newcommand\yGyroBar[2]{{\prescript{{#2}}{}{\bar{\omega}}_{#1, #2}}}
\newcommand\biasGyro{{{\mathbf{b}}_{g}}}
\newcommand\biasGyroHat{{\mathbf{\hat{b}}_g}}
\newcommand\noiseGyro[1]{{\mathbf{w}_{g}^{#1}}}

\newcommand\noiseLinVel[1]{{\mathbf{w}_{v}^{#1}}}
\newcommand\noiseAngVel[1]{{\mathbf{w}_{\omega}^{#1}}}

\newcommand\fkNoiseLinVel[1]{{\mathbf{n}_{v}^{#1}}}
\newcommand\fkNoiseAngVel[1]{{\mathbf{n}_{\omega}^{#1}}}

\newcommand\encoders{\mathbf{\tilde{s}}}

\newcommand\encoderNoise{\mathbf{w}_\mathbf{s}}

\newcommand\TransformMeasured[2]{{\prescript{{#1}}{}{\mathbf{\tilde{H}}}_{#2}}}

\newcommand{\cov}{\mathbf{P}}
\newcommand{\Q}{\mathbf{Q}}
\newcommand{\N}{\mathbf{N}}

\newcommand{\F}{\mathbf{F}}

\newcommand{\Xhat}{\mathbf{\hat{X}}}
\newcommand{\err}{\epsilon}
\newcommand{\errG}{\eta}

\newcommand\kpred{{k+1 \mid k}}
\newcommand\kprior{{k \mid k}}
\newcommand\kest{{k+1 \mid k+1}}
\newcommand\kcurr{{k}}
\newcommand\knext{{k+1}}
\newcommand\knextgivenl{{k+1 \mid l}}

\newcommand{\expectation}[1]{\mathbb{E}\hm{[} {#1} \hm{]}}

\newcommand\gravity[1]{{\prescript{{#1}}{}{\mathbf{g}}}}
\newcommand\Transform[2]{{\prescript{{#1}}{}{\mathbf{H}}_{#2}}}

\newcommand\Rot[2]{{\prescript{{#1}}{}{\mathbf{R}}_{#2}}}
\newcommand\Pos[2]{{\prescript{{#1}}{}{\mathbf{o}}_{#2}}}
\newcommand\PosBar[2]{{\prescript{{#1}}{}{\mathbf{\bar{o}}}_{#2}}}
\newcommand\oDot[2]{{\prescript{{#1}}{}{\mathbf{\dot{o}}}_{#2}}}
\newcommand\oDoubleDot[2]{{\prescript{{#1}}{}{\mathbf{\ddot{o}}}_{#2}}}

\newcommand\relativeJacobianLeftTriv[2]{{\prescript{{#2}}{}{\mathbf{S}}_{#1, #2}}}

\newcommand\twistMixedTriv[2]{{\prescript{{#2[#1]}}{}{\textbf{v}}_{#1, #2}}}

\newcommand\omegaLeftTriv[2]{{\prescript{{#2}}{}{\omega}_{#1, #2}}}
\newcommand\omegaRightTriv[2]{{\prescript{{#1}}{}{\omega}_{#1, #2}}}

\newcommand\jointPos{\mathbf{{s}}}

\title{\LARGE \bf
An Experimental Comparison of Floating Base Estimators for Humanoid Robots with Flat Feet
}

\author{Prashanth Ramadoss$^{1, 2}$, Stefano Dafarra$^{1}$,  Silvio Traversaro$^{1}$, and Daniele Pucci$^{1}$ 
\thanks{$^{1}$ Artificial and Mechanical Intelligence, Italian Institute of Technology,
Genoa, Italy, {\tt\small (e-mail: name.surname@iit.it)}}
\thanks{$^{2}$ DIBRIS, University of Genoa, Genoa, Italy}
}

\begin{document}

\maketitle
\thispagestyle{empty}
\pagestyle{empty}


\begin{abstract}
Extended Kalman filtering is a common approach to achieve floating base estimation of a humanoid robot.
These filters rely on measurements from an Inertial Measurement Unit (IMU) and relative forward kinematics for estimating the base position-and-orientation and its linear velocity along with the augmented states of feet position-and-orientation, thus giving them their name,  \emph{flat-foot filters}.
However, the availability of only partial measurements often poses the question of consistency in the filter design.
In this paper, we perform an experimental comparison of state-of-the-art flat-foot filters based on the representation choice of state, observation, matrix Lie group error and system dynamics evaluated for filter consistency and trajectory errors.
The comparison is performed over simulated and real-world experiments conducted on the iCub humanoid platform.
\looseness=-1
\end{abstract}

\section{Introduction}
\par
A commonly used approach for floating base estimation in the humanoid robots' community relies on extended Kalman filtering based \emph{flat-foot filters}.
This  paper  contributes  towards  the comparison of flat-foot filters while trying to investigate a few relevant modeling choices helpful for such a filter design applied to kinematic-inertial odometry.

An Observability-Constrained EKF (OCEKF) was proposed by \cite{bloesch2013state} for a consistent fusion of IMU and encoder measurements along with contact states for the base estimation of a quadrupedal robot.
This approach considered the augmentation of the base state with feet positions through point foot contact models.
OCEKF was then extended by \cite{rotella2014state} for the flat-foot contact scenario of humanoid robots which also incorporated the feet rotations within the state leading to an improved accuracy in estimation.
Recently, \cite{hartley2020contact} proposed an Invariant Extended Kalman Filtering (InvEKF) using the theory of matrix Lie groups exhibiting strong convergence and consistency properties.
This filter exploits a $\SEk{k+2}{3}$ matrix Lie group introduced by \cite{barrau2017invariant} for state representation along with the group-affine dynamics property enabling autonomous error propagation and invariant measurement updates, leading to an error evolution independent of the state trajectory.
InvEKF has been extended to the flat-foot scenario of the humanoid robots (InvEKF-F) by \cite{qin2020novel} to consider the foot rotations within the state.
A similar extension, named as DILIGENT-KIO, using a discrete Lie group extended Kalman filter considering the evolution of both the states and measurements on distinct matrix Lie groups has been developed in \cite{ramadoss2021diligent} using a state representation different from the one used by \cite{qin2020novel, hartley2020contact}. \looseness=-1

A significant effort has been made towards the formulation of \emph{consistent} state estimation in the SLAM community (\cite{huang2010observability, barrau2017invariant}), and these are being actively applied to humanoid base estimation (\cite{rotella2014state, qin2020novel}).
A consistent estimator is characterized by zero-mean estimation errors with covariance smaller than or equal to that estimated by the filter.
When designing an estimator with partial observations, as in the case of flat-foot filters, a proper covariance management is crucial for maintaining the consistency of the filter.
The availability of only partial measurements may lead to the existence of a few unobservable directions in the underlying nonlinear system. 
If the linearized error system becomes dependent on the state configuration, then a wrong choice of linearization point may cause the state covariance matrix to gain information in the unobservable directions, leading to overconfident estimates.
This may, in turn, cause an inconsistency in the filter which could result in a divergence of the estimated states. \looseness=-1

As the practitioner, we pose the following questions towards reliable filter design: \emph{1) how does the representation choice for state and observations affect the filter design? 2) how does the filter design vary with the choice of error? 3) how does the time-representation of the system dynamics affect the design?}
Based on these choices, the linearized error system may or may not be independent of the trajectory of the system state.
Further, a well-chosen matrix Lie group representation  allows for a design \cite{barrau2017invariant} with which, a wide range of nonlinear problems can lead to linear error equations through a correct parameterization of the error. 
Proper choices for these questions often helps in formulating an estimator with strong convergence and consistency properties. \looseness=-1

In this paper, we address the questions posed above through an experimental evaluation of state-of-the-art flat foot filters, see Table \ref{table:soa}.
In order to aid the comparison, we derive a few variations of our previously proposed estimator, DILIGENT-KIO \cite{ramadoss2021diligent}, based on the choice of matrix Lie group error and time-representation for the system dynamics.
The contributions of this paper are as follows, \looseness=-1
\begin{itemize}
    \item Development of DILIGENT-KIO with a right-invariant error formulation (DILIGENT-KIO-RIE), contrary to its original counterpart which uses a left-invariant error, and their subsequent modifications to use continuous system dynamics leading to CODILIGENT-KIO and CODILIGENT-KIO-RIE;
    \item Open-source implementation \footnote{\url{https://github.com/ami-iit/paper_ramadoss-2022-ral-humanoid-base-estimation}} and evaluation of the flat-foot filters for experiments conducted in simulation and on the real robot using iCub humanoid platform. \looseness=-1
\end{itemize}

This paper is organized as follows. Section \ref{sec:BACKGROUND} introduces the mathematical concepts related to filtering on matrix Lie groups. 
Section \ref{sec:ESTIMATION} describes the estimator design approach used for the derivation of DILIGENT-KIO variants and subsequently Section \ref{sec:FILTERS} briefly describes relevant Jacobian computations for these filters.
This is followed by experimental evaluation of the flat-foot filters from the state-of-the-art in Section \ref{sec:RESULTS} and concluding remarks in Section \ref{sec:CONCLUSION}.

\vspace{1mm}
\begin{table}[t]
\caption{Comparison of contact-aided kinematic-inertial flat foot filters.}
\centering
\scalebox{0.9}{
\begin{tabular}{p{3cm} |  p{1.2cm} | p{0.8cm}| p{1.3cm}| p{1.2cm} }
\hline \rowcolor[gray]{.9}
Author, Year  &       Autonomous Error Propagation     &   Invariant Observations    &    Observability Constraints &  Filter Design on Matrix Lie Groups         \\
\hline
Rotella, 2014 (OCEKF) \cite{rotella2014state} & \ding{55} & \ding{55} & \checkmark &  \ding{55} \\
\rowcolor[gray]{.9} Ramadoss, 2021 (DILIGENT-KIO) \cite{ramadoss2021diligent}     &   \ding{55}     &   \ding{55}    &  \ding{55} & \checkmark \\
Ramadoss, 2022 (DILIGENT-KIO-RIE)   &   \ding{55}   &  \ding{55}   &  \ding{55} & \checkmark \\
\rowcolor[gray]{.9} Ramadoss, 2022 (CODILIGENT-KIO) &   \checkmark     &   \ding{55}    &        \ding{55} & \checkmark \\
Ramadoss, 2022 (CODILIGENT-KIO-RIE)      &   \checkmark     &  \ding{55}    &            \ding{55} & \checkmark \\
\rowcolor[gray]{.9} Qin, 2020 (InvEKF-F)  \cite{qin2020novel} &   \checkmark     &   \checkmark    &  not needed & \checkmark \\
\hline
\end{tabular}
}
\label{table:soa} 
\vspace{-3mm}
\end{table}
\rowcolors{0}{}{}
\section{Mathematical Background}
\label{sec:BACKGROUND}

\par

\subsection{Notations and definitions}
\subsubsection{Coordinate Systems}
\begin{itemize}
	
	\item ${C[D]}$ denotes a frame with origin $\Pos{}{C}$ and orientation of coordinate frame ${D}$;
	\item $\Pos{A}{B}\in \R^3$ and $\Rot{A}{B}\in \SO{3}$ are the position and orientation of a frame $B$ with respect to the frame $A$;
	
	\item given $\Pos{A}{C}$ and $\Pos{B}{C}$,  $\Pos{A}{C} = \Rot{A}{B} \Pos{B}{C} + \Pos{A}{B}= \Transform{A}{B} \PosBar{B}{C}$, where $\Transform{A}{B} \in \SE{3}$ is the homogeneous transformation and $\PosBar{B}{C} = \begin{bmatrix}\Pos{B}{C}^T & \ 1\end{bmatrix}^T \in \R^4$ is the homogeneous representation of $\Pos{B}{C}$; 
	
	\item $\twistMixedTriv{A}{B} = \oDot{A}{B} = \frac{d}{dt}(\Pos{A}{B}) \in \R^3$ denotes the linear part of a mixed-trivialized velocity \cite[Section 5.2]{traversaro2019multibody} between the frame $B$ and the frame $A$ expressed in the frame ${B[A]}$. $\omegaRightTriv{A}{B} \in \R^3$ denotes the angular velocity between the frame ${B}$ and the frame ${A}$ expressed in ${A}$; \looseness=-1
	\item ${A}$ denotes an absolute or an inertial frame; ${B}$, ${LF}$, ${RF}$ and ${S}$ indicate the frames attached to the base link, left foot, right foot and the IMU rigidly attached to the base link respectively; \looseness=-1
\end{itemize}	
\subsubsection{Lie Groups}

\begin{itemize}
\item $G, G^\prime \subset \R^{n \times n}$ denote matrix Lie groups and $\X, \Y \in G$ are elements of the matrix Lie group $G$. 
\item $\mathfrak{g, g^\prime} \subset \R^{n \times n}$ denote the matrix Lie algebras for the groups $G, G^\prime$ respectively.
\item $\ghat{G}{.}: \R^p \to \mathfrak{g}$ and $\gvee{G}{.}: \mathfrak{g} \to \R^p$ are the \emph{hat} and \emph{vee} operators for the matrix Lie group $G$ which denote a linear isomorphism between $\mathfrak{g}$ and a $p$-dimensional vector space. $\forall \;\mathfrak{a} \in \mathfrak{g}$, \; $\aVec = \gvee{G}{\mathfrak{a}} \in \R^p$, $\mathfrak{a} = \ghat{G}{\aVec}$.
\item $\gexphat{G}: \R^p \to G$ is the exponential map operator that maps elements from the vector space associated to the Lie algebra directly to elements of the group. $\forall \aVec \in \R^p, \; \gexphat{G}(\aVec) = \gexp{G}(\ghat{G}{\aVec})$.
\item $\glogvee{G}: G \to \R^p$ is the logarithm map operator that maps elements of the group directly to the vector space associated to the Lie algebra. $\forall \; \X \in G, \; \glogvee{G}(\X) = \glog{G}(\gvee{G}{\X})$. This mapping may not be unique. \looseness =-1
\item $\gadj{}: \R^p \to \R^p$ is the adjoint matrix operator that linearly transforms vectors of the tangent space at an element $\X$ onto the Lie algebra and can be computed as, $\forall \; \X \in G, \aVec \in \R^p, \mathfrak{a} \in \mathfrak{g}, \;  \gadj{\X}\;\aVec = \gvee{G}{\X \mathfrak{a} \X^{-1}}$.  \looseness =-1
\item $\forall \; \aVec \in \R^p, \; \grjac{G}(-\aVec)$ denotes the right Jacobian of the matrix Lie group that relates any perturbations in the parametrizations of the Lie group to the changes in the group velocities $\X^{-1} \dot{\X}$. Similarly, the left Jacobian $\gljac{G}(\aVec)$ relates those to the changes in $\dot{\X}\X^{-1}$ \cite{chirikjian2011stochastic}.
\end{itemize}

\subsubsection{Miscellaneous}
\begin{itemize}
\item $\I{n}$ and $\Zero{n}$ denote the $n \times n$ identity and zero matrices;

\item given $\uVec, \vVec \in \R^3$ the \emph{hat operator} for $\SO{3}$ is $S(.): \R^3 \to \so{3}$, where $\so{3}$ is the set of skew-symmetric matrices and $S(\uVec) \; \vVec = \uVec \times \vVec$;  $\times$ is the cross product operator in $\R^3$. \looseness=-1

\end{itemize}

\subsection{Invariant errors}
For matrix Lie groups, there is a possibility of constructing \emph{invariant errors} that remain the same even for a system transformed under a left- or right-translation. 

\begin{definition}[Left and right invariant error of type 1]
For $\mathbf{S}, \Y \in G$, given the definition of left translation $L_\mathbf{S}: G \to G, \ L_\mathbf{S} \Y = \mathbf{S} \Y$ and the right translation $R_\mathbf{S}: G \to G, \ R_\mathbf{S} \Y = \Y \mathbf{S}$, the invariant errors between a true state $\X \in G$ and an estimated state $\Xhat \in G$ are, \looseness=-1
\begin{align}
\label{def:left-invariant-error}
&\errG^L = \Xhat^{-1}\ \X = (L_\mathbf{S}\Xhat )^{-1} (L_\mathbf{S} \X ) \quad \text{(left-invariant)}, \\
\label{def:right-invariant-error}
&\errG^R = \X\ \Xhat^{-1} = (R_\mathbf{S}\X ) (R_\mathbf{S} \Xhat )^{-1} \quad \text{(right-invariant)}.
\end{align}
\end{definition}

The invariant errors can also be defined as a type 2, $\errG^L = \X^{-1}\ \Xhat$ and $\errG^R = \Xhat\ \X^{-1}$.

\begin{remark}
The invariant errors considered for the derivations in this article are of type 1 (Eqs. \eqref{def:left-invariant-error} and \eqref{def:right-invariant-error}).
\end{remark}

\subsection{Uncertainty on matrix Lie groups}

The concept of Concentrated Gaussian Distribution (CGD) is used to define the notion of uncertainty for the matrix Lie groups \cite{bourmaud2015continuous, barfoot2014associating} producing a distribution on $G$ centered at $\Xhat$,\looseness=-1
\begin{align}
\small
 \X = \Xhat \gexphat{G}(\err), \quad \text{(perturbations $\err$ applied locally)}, \\ 
 \small
 \X = \gexphat{G}(\err) \Xhat,   \quad \text{(perturbations $\err$ applied globally)},
\end{align}
where, $\Xhat \in G \subset \R^{n \times n}$ is the mean of $\X$ defined on the group and $\epsilon \in \R^p$ is a small perturbation having a zero-mean Gaussian distribution defined in the $p$-dimensional vector space associated to the state, with the covariance matrix $\cov \in \R^{p \times p}$ defined in the Lie algebra.


\subsection{Discrete extended Kalman filter on matrix Lie groups}
Consider a discrete dynamical system for which both the state and the  measurements are evolving over distinct matrix Lie groups $G$ and $G^\prime$ respectively, \looseness=-1
\begin{align}
\label{eq:sec:background:ekf-lie-group-system}
\X_\knext &=  \X_\kcurr \; \gexphat{G}\left(\Omega\left(\X_\kcurr, \mathbf{u}_\kcurr\right) + \mathbf{w}_\kcurr\right), \\
\label{eq:sec:background:ekf-lie-group-meas}
\Z_\kcurr &= h (\X_\kcurr) \; \gexphat{G^\prime}\left(\mathbf{n}_\kcurr\right).
\end{align}
\noindent The function $\Omega : G \times \R^m \to \R^p$ is a left trivialized velocity expressed as a function of the state $\X_\kcurr \in G$ and an exogenous control input $\mathbf{u}_\kcurr \in \R^m$ at time instant $\kcurr$.
We use $\mathbf{w}_\kcurr\sim \mathcal{N}_{\R^p}(\mathbf{0}_{p \times 1},  \Q_\kcurr)$ to denote discrete-time Gaussian white noise with covariance $\Q_\kcurr \in \R^{p \times p}$ acting on the motion model. \looseness=-1

The observations $\Z_\kcurr$ are considered to be evolving over a matrix Lie group $G^\prime$ of dimensions $q$. 
The function $h: G \to G^\prime$ is the measurement model mapping the states $\X \in G$ to the space of observations $G^\prime$. We use $\mathbf{n}_\kcurr\sim\mathcal{N}_{\R^{q}}(\mathbf{0}_{q\times 1}, \; \N_\kcurr)$ to denote the measurement noise described as a discrete-time Gaussian white noise with covariance $\N_\kcurr$ defined in the $q$-dimensional vector space of the observations.\looseness=-1

The filtering problem, then, involves finding the optimal estimate $\Xhat$ given a set of observations ${\Z_1,\dots, \Z_l}$ at time instants $l = \kcurr$ for the propagation and $l = \knext$ for the update.
The estimated state at each step is given by the mean $\Xhat_\knextgivenl$ of the conditional probability and its covariance is $\cov_\knextgivenl$. \looseness=-1
\begin{align}
\label{eq:sec:background:ekf-posterior-distribution}
\begin{split}
    &p(\X_\knext\; \lvert\; \Z_1 \dots \Z_l) \approx \mathcal{N}_G(\Xhat_\knextgivenl, \;\cov_\knextgivenl), \\
    &\X_\knextgivenl = \Xhat_\knextgivenl\; \gexphat{G}({\err_\knextgivenl}), \\
    &\err_\knextgivenl \sim \mathcal{N}_{\R^p}(\mathbf{m}_\knextgivenl = \mathbf{0}_{p \times 1}, \cov_\knextgivenl), 
\end{split}
\end{align}
where, $p(\X_\knext\; \lvert\; \Z_1 \dots \Z_l)$ is the conditional probability distribution characterized as a concentrated Gaussian distribution. 

If the mean and covariance of the CGD are propagated and updated considering  the left-invariant error formulation of type 1 denoted as $\errG = \errG^L = \Xhat^{-1} \X$ (see Eq. \eqref{def:left-invariant-error}), then
the resulting estimator design leads to the \emph{Discrete Lie Group Extended Kalman Filter} (abbreviated as DILIGENT) whose equations are summarized in Algorithm \ref{algo:sec:background:dlgekf}.

On considering  the right-invariant error formulation of type 1 denoted as $\errG = \errG^R = \X \Xhat^{-1}$ (see Eq. \eqref{def:right-invariant-error}), the resulting filter design is the \emph{Discrete Lie Group Extended Kalman Filter with Right Invariant Error} (abbreviated as DILIGENT-RIE) whose equations are summarized in Algorithm \ref{algo:sec:background:dlgekf-rie}.
When the right-invariant error is chosen, the error propagation, update and the state reparametrization are different from the left-invariant error variant accordingly.
\scalebox{0.73}{
\parbox{.5\linewidth}{
\begin{algorithm}[H]
\small
  \caption{Discrete extended Kalman filter on matrix Lie groups with \textbf{left-invariant error} formulation of type 1 \looseness=-1}
  \label{algo:sec:background:dlgekf}
  \begin{algorithmic}
    \State \textbf{Input:} $\Xhat_\kprior, \cov_\kprior, \Z_\knext, \mathbf{u}_\kcurr$ \\ 
    \textbf{Output:} $\Xhat_\kest, \cov_\kest$ \\ \hrulefill
    \State \textbf{Propagation:}\\
    $\Xhat_\kpred = \Xhat_\kprior\; \gexphat{G}\left(\hat{\Omega}_\kcurr\right)$ \\
    $\cov_\kpred =\ \F_\kcurr\ \cov_\kprior\ \F_\kcurr^T\ +\ \grjac{G}\left(\hat{\Omega}_\kcurr\right)\ \Q_\kcurr\ \grjac{G}\left(\hat{\Omega}_\kcurr\right)^T$ \\ \hrulefill
    \State \textbf{Update:} \\
    $\mathbf{\tilde{z}}_\knext =  \glogvee{G^\prime}\left( h^{-1}\left(\Xhat_\kpred\right)\ \Z_\knext\right)$ \\
    $\mathbf{K}_\knext = \cov_\kpred\ \mathbf{H}_\knext^T\left(\mathbf{H}_\knext\;\cov_\kpred\;\mathbf{H}_\knext^T \;+\; \mathbf{N}_\knext\right)^{-1}$ \\
    $\mathbf{m}_\knext^{-} = \mathbf{K}_\knext\; \mathbf{\tilde{z}}_\knext$ \\
    $\Xhat_\kest\ =\ \Xhat_\kpred \gexphat{G}\left(\mathbf{m}_\knext^{-}\right)$ \\
    $\cov_\kest =  \grjac{G}\left(\mathbf{m}_\knext^{-}\right) \left(\I{p} \;-\; \mathbf{K}_\knext\;\mathbf{H}_\knext\right)\;\cov_\kpred\;  \grjac{G}\left(\mathbf{m}_\knext^{-}\right)^T$\\ \hrulefill

    \State where, \\
    $\hat{\Omega}_\kcurr = \Omega\left(\Xhat_\kprior, \mathbf{u}_\kcurr\right)$ \\
    $\F_\kcurr\ = \gadj{\gexphat{G}\left(-\hat{\Omega}_\kcurr\right)} + \grjac{G}\left(\hat{\Omega}_\kcurr\right)\ \mathfrak{F}_\kcurr$ \\
    $\mathfrak{F}_\kcurr = \frac{\partial}{\partial \err} \Omega\left(\Xhat_\kprior\ \gexphat{G}\left({\err_\kprior}\right), \mathbf{u}_\kcurr\right)_{\err_\kprior = 0}$ \\
    $\mathbf{H}_\knext = \frac{\partial}{\partial \err} \glogvee{G^\prime}\left( h^{-1}\left(\Xhat_\kpred\right)\ h\left(\Xhat_\kpred\ \gexphat{G}\left(\err_\kpred\right)\right)\right)_{\err=0}$ \\ \\
    \vspace{-1mm}
  \end{algorithmic}
  \vspace{-2mm}
\end{algorithm}
}
} 

\scalebox{0.75}{
\parbox{.5\linewidth}{
\begin{algorithm}[H]
\small
  \caption{Discrete extended Kalman filter on matrix Lie groups with \textbf{right-invariant error} formulation of type 1 \looseness=-1}
  \label{algo:sec:background:dlgekf-rie}
  \begin{algorithmic}
    \State \textbf{Input:} $\Xhat_\kprior, \cov_\kprior, \Z_\knext, \mathbf{u}_\kcurr$ \\ 
    \textbf{Output:} $\Xhat_\kest, \cov_\kest$ \\ \hrulefill
    \State \textbf{Propagation:}\\
    $\Xhat_\kpred = \Xhat_\kprior\; \gexphat{G}\left(\hat{\Omega}_\kcurr\right)$ \\
    \scalebox{0.95}[1]{$\cov_\kpred =\ \F_\kcurr\ \cov_\kprior\ \F_\kcurr^T\ +\ \gadj{\Xhat_\kprior} \gljac{G}\left(\hat{\Omega}_\kcurr\right)\ \Q_\kcurr\ \gljac{G}\left(\hat{\Omega}_\kcurr\right)^T \gadj{\Xhat_\kprior}^T$}\\ \hrulefill
    \State \textbf{Update:} \\
    $\mathbf{\tilde{z}}_\knext =  \glogvee{G^\prime}\left( h^{-1}\left(\Xhat_\kpred\right)\ \Z_\knext\right)$ \\
    $\mathbf{K}_\knext = \cov_\kpred\ \mathbf{H}_\knext^T\left(\mathbf{H}_\knext\;\cov_\kpred\;\mathbf{H}_\knext^T \;+\; \mathbf{N}_\knext\right)^{-1}$ \\
    $\mathbf{m}_\knext^{-} = \mathbf{K}_\knext\; \mathbf{\tilde{z}}_\knext$ \\
    $\Xhat_\kest\ =\ \gexphat{G}\left(\mathbf{m}_\knext^{-}\right) \Xhat_\kpred$   \\
    $\cov_\kest =  \gljac{G}\left(\mathbf{m}_\knext^{-}\right) \left(\I{p} \;-\; \mathbf{K}_\knext\;\mathbf{H}_\knext\right)\;\cov_\kpred\;  \gljac{G}\left(\mathbf{m}_\knext^{-}\right)^T$\\ \hrulefill

    \State where, \\
    $\hat{\Omega}_\kcurr = \Omega\left(\Xhat_\kprior, \mathbf{u}_\kcurr\right)$ \\
    $\F_\kcurr\ = \I{p} + \gadj{\Xhat_\kprior} \gljac{G}\left(\hat{\Omega}_\kcurr\right)\ \mathfrak{F}_\kcurr$ \\
    $\mathfrak{F}_\kcurr = \frac{\partial}{\partial \err} \Omega\left(\gexphat{G}\left({\err_\kprior}\right) \Xhat_\kprior, \mathbf{u}_\kcurr\right)_{\err_\kprior = 0}$ \\
    $\mathbf{H}_\knext = \frac{\partial}{\partial \err} \glogvee{G^\prime}\left( h^{-1}\left(\Xhat_\kpred\right)\ h\left(\gexphat{G}\left(\err_\kpred\right) \Xhat_\kpred\ \right)\right)_{\err=0}$ \\ \\

    \vspace{-1mm}
  \end{algorithmic}
  \vspace{-2mm}
\end{algorithm}
}
} 

\subsection{Autonomous Error Propagation}
A particular class of continuous-time systems can often lead to some desirable error-propagation properties for the filter design, which can then be discretized for discrete-time implementation.
Consider a continuous-time system evolving over matrix Lie groups given by, \looseness=-1
\begin{align}
\label{eq:sec:background:invekf-dynamical-system}
\frac{d}{dt} \X_t = f_{\mathbf{u}_t}\left(\X_t\right),
\end{align}
where, the state evolves over the Lie group $\X_t \in G$, $\mathbf{u}_t \in \R^m$ is the exogenous control input and the system dynamics is given by $f_{\mathbf{u}_t}\left(\X_t\right) \triangleq f\left(\X_t, \mathbf{u}_t\right)$.
If the system follows a \emph{group-affine} dynamics, then the estimation error obeys an autonomous equation, i.e., the error propagation is independent of the state trajectory.
This is given by the theorem of \emph{autonomous error dynamics} \cite[Theroem 1]{barrau2017invariant}.

\begin{definition}[Autonomous error \cite{barrau2017invariant}]
\label{def:sec:background:automonous-error}
The invariant errors have a state-trajectory independent propagation if they satisfy a differential equation of the form $\frac{d}{dt} \errG_t = g_{\mathbf{u}_t}\left(\errG_t\right)$.
\end{definition}

\begin{definition}[Group-affine dynamics \cite{barrau2017invariant}]
\label{def:sec:background:group-affin-dyn}
System dynamics is said to be group-affine, if $\forall t > 0$ and $\X_1, \X_2 \in G$, 
 \begin{equation*}
    \label{eq:sec:background:invekf-group-affine-dynamics}
        f_{\mathbf{u}_t}\left(\X_1\ \X_2\right) = f_{\mathbf{u}_t}\left(\X_1\right) \X_2 +  \X_1\ f_{\mathbf{u}_t}\left(\X_2\right) - \X_1 f_{\mathbf{u}_t}\left(\I{p}\right) \X_2.
\end{equation*}
\end{definition}

Assuming the error to be small ($\errG_t = \gexphat{G}\left(\err_t\right) \approx \I{n}$), the results of \cite[Theroem 1]{barrau2017invariant} can be extended to a noisy model of the form, \looseness=-1
\begin{align}
\label{eq:sec:background:invekf-noisy-dynamical-system}
\frac{d}{dt} \X_t = f_{\mathbf{u}_t}\left(\X_t\right) + \X_t\ \ghat{G}{\mathbf{w}_t},
\end{align}
where, $\mathbf{w}_t \sim \mathcal{N}_{\R^p}(\mathbf{0}_{p \times 1},  \Q_t)$ is a continuous-time white noise belonging to $\mathfrak{g}$ with covariance $\Q_t$ defined in $\R^p$.
The left- and right-invariant error propagation can be written as,
\begin{align}
    & \frac{d}{dt} \errG_t^L = g^L_{\mathbf{u}_t}(\errG_t^L) - \ghat{G}{\mathbf{w}_t}\ \errG^L_t, \\
    & \frac{d}{dt} \errG_t^R = g^R_{\mathbf{u}_t}(\errG_t^R) - \left(\Xhat_t\ \ghat{G}{\mathbf{w}_t}\ \Xhat_t^{-1}\right) \errG^R_t,
\end{align}
and with the definition $\A_t \in \R^{p \times p}, \ g_{\mathbf{u}_t}(\gexphat{G}(\err_t)) = \gexphat{G}(\A_t \err_t) + \mathcal{O}(\norm{\err}^2)$, the linearized error equations become,
\begin{align}
    \label{eq:sec:background:invekf-linearized-lie-dynamics}
    & \frac{d}{dt} \err^L_t = \A^L_{t}\ \err^L_t -  \mathbf{w}_t, \\
    \label{eq:sec:background:invekf-linearized-rie-dynamics}
    & \frac{d}{dt} \err^R_t = \A^R_{t}\ \err^R_t - \gadj{\Xhat_t} \mathbf{w}_t.
\end{align}
The linearized error equations lead to covariance propagation, 
\begin{align}
    \begin{split}
        & \frac{d}{dt}\cov_t = \mathbf{A}_{t} \cov_t +\ \cov_t \mathbf{A}_{t}^T\ +\ \mathbf{\hat{Q}}_t,
    \end{split}
\end{align}
where, $\mathbf{\hat{Q}}_t$ is the modified process noise and measurement noise covariance matrices depending on the choice of the invariant error.
The autonomous error and log-linear error equations play a significant role in the \emph{invariant EKF} design leading to state-independent linearized error systems.

\emph{Continuous-Discrete Lie Group Extended Kalman Filter} (CODILIGENT) with a left-invariant error and its right-invariant error counterpart CODILIGENT-RIE can then be designed for systems of the form, \looseness=-1
\begin{equation}
    \label{eq:sec:background:cdekf-rie}
    \begin{split}
    & \frac{d}{dt}\X_t = f_{\mathbf{u}_t}\left(\X_t\right) + \X_t \ghat{G}{\mathbf{w}_t}, \\
    & \Z_{t_\kcurr} = h (\X_{t_\kcurr}) \; \gexphat{G^\prime}(\mathbf{n}_{t_\kcurr}).
    \end{split}
\end{equation}
It must be noted that this filter design is different from the one proposed in \cite{bourmaud2015continuous}, where the latter follows a more rigorous treatment of continuous system dynamics.
Further, this design does not exploit the property of \emph{invariant observations} \cite{barrau2017invariant}.

\section{Estimator Design for Kinematic-Inertial Odometry (KIO)}
\label{sec:ESTIMATION}

\subsection{Problem Description}
\begin{assumption}
\label{assumption:sec:estimation:diligent-kio-imu-collocation}
An IMU is rigidly attached to the base link and their coordinate frames coincide.
\end{assumption}
\begin{assumption}
\label{assumption:sec:estimation:diligent-kio-rigid-foot-contact}
At least one foot of the robot is always in rigid contact with the environment.
\end{assumption}
For floating base estimation of the humanoid robot, we wish to estimate the position $\Pos{A}{B}$, orientation $\Rot{A}{B}$ and linear velocity $\oDot{A}{B}$ of the base link in the inertial frame, written in shorthand as $\mathbf{p}, \mathbf{R}$, and $\mathbf{v}$ respectively. \looseness=-1
Additionally, we augment the state with position $\Pos{A}{F}$ and orientation $\Rot{A}{F}$ of the foot link in the inertial frame, in shorthand $\mathbf{d}_F$ and $\mathbf{Z}_F$,
where, $F = \{ {LF}, {RF} \}$ is the set containing left foot and right foot frames.
Since we also rely on the base-collocated IMU for the state estimation, it is necessary to also estimate the slowly time-varying biases $\biasAcc$ and $\biasGyro$ affecting the accelerometer and gyroscope measurements, respectively.

\subsection{State Representation}
The tuple representation of a matrix $ \X_B =  (\mathbf{p}, \mathbf{R}, \mathbf{v})_{\SEk{2}{3}}$ represent the base link quantities, while the feet quantities are represented as $\X_F = (\mathbf{d}_F, \mathbf{Z}_F)_{\SE{3}}$.
The biases are sometimes included in a single vector as $\bias = \begin{bmatrix}\biasAcc^T & \biasGyro^T\end{bmatrix}^T$. \looseness=-1
A matrix Lie group element from the state space $\mathcal{M}$ is then, 
\setcounter{MaxMatrixCols}{20}
\begin{align}
\label{eq:sec:estimation:diligent-kio-state}
\begin{split}
&\X = \text{blkdiag}(\X_B, \X_{LF}, \X_{RF}, \X_{\bias}) \in \mathcal{M} \subset \R^{20 \times 20}, 
\end{split}
\end{align}
where, the matrix sub-blocks are given by,
\begin{align*}
\begin{split}
& {\footnotesize \X_B =  \begin{bmatrix} 
\Rot{}{} & \mathbf{p} & \vVec \\ \Zeros{1}{3} & 1 & 0 \\ \Zeros{1}{3} & 0 & 1\end{bmatrix}}, \X_F = {\footnotesize \begin{bmatrix}
\Z_F & \mathbf{d}  \\ \Zeros{1}{3} & 1
\end{bmatrix}},  \X_\bias = {\footnotesize \begin{bmatrix}
\I{6} & \bias  \\ \Zeros{1}{6} & 1
\end{bmatrix}}.
\end{split}
\end{align*}
For convenience, a tuple representation can be used as a short-hand notation for an element from the state space, \looseness=-1
$$
\X = \left(\mathbf{p}, \mathbf{R}, \mathbf{v}, \mathbf{d}_{LF}, \mathbf{Z}_{LF}, \mathbf{d}_{RF}, \mathbf{Z}_{RF}, \bias \right)_{\mathcal{M}}.
$$
The vector $\epsilon \in \R^{27}$ associated with the Lie algebra is, 
\begin{align}
\label{eq:sec:estimation:diligent-kio-state-vector}
\begin{split}
\epsilon &= \text{vec}(\epsilon_B, \epsilon_{LF}, \epsilon_{RF}, \epsilon_\bias)  \\
& = \text{vec}\left(\epsilon_\mathbf{p},\ \epsilon_\mathbf{R},\ \epsilon_\mathbf{v},\ \epsilon_{\mathbf{d}_{LF}},\ \epsilon_{\mathbf{Z}_{LF}},\ \epsilon_{\mathbf{d}_{RF}},\ \epsilon_{\mathbf{Z}_{RF}}, \epsilon_\bias \right).
\end{split}
\end{align}
The Lie group operator definitions for the state representation can be found in \cite{ramadoss2021diligent}.
 \looseness=-1 

\subsection{System Dynamics}
A  strap-down IMU-based kinematics model is used for tracking the base state. 
This model uses accelerometer and gyroscope measurements $(\yAcc{A}{B}, \yGyro{A}{B})$ as exogenous inputs that are modeled to be affected by slowly time-varying biases $\biasAcc$ and $\biasGyro$, respectively, along with additive white Gaussian noise $\noiseAcc{B}$ and $\noiseGyro{B}$, as seen in Eq. \eqref{eq:sec:estimation:diligent-kio-imu-sensor-model}. 
\begin{align}
\label{eq:sec:estimation:diligent-kio-imu-sensor-model}
\begin{split}
&\yAcc{A}{B} = {\Rot{}{}}^T (\oDoubleDot{A}{B} - \gravity{A}) + \biasAcc + \noiseAcc{B} \\
&\yGyro{A}{B} = \omegaLeftTriv{A}{B} + \biasGyro + \noiseGyro{B}
\end{split}
\end{align}

With the Assumption \ref{assumption:sec:estimation:diligent-kio-rigid-foot-contact}, holonomic constraints are imposed whenever the foot makes contact with the environment causing the foot velocity to be zero. 
The noises $\noiseLinVel{F}$ and $\noiseAngVel{F}$ affecting null foot velocities are expressed locally in the foot frame.
A null velocity model is considered for the foot pose to respect the contact events. 
But, this motion model becomes invalid when the foot is in the swing phase, thus the variances related to the foot velocities are dynamically scaled to very high values causing the estimated foot pose from the prediction model to grow uncertain.
The measurement updates are then relied upon to update the foot pose causing it to reset to a more reliable estimate, whenever a contact is made. \looseness=-1

The overall continuous system dynamics then take the form, \looseness=-1
\begin{equation}
    \label{eq:sec:estimation:cdekf-sys-dynamics}
    \begin{split}
    & \dot{\mathbf{{p}}} = \mathbf{v},    \\
    & \dot{\mathbf{\Rot{}{}}} = \Rot{}{}\ S(\yGyro{A}{B} - \biasGyro - \noiseGyro{B}), \\
    & \dot{\mathbf{{v}}} = \Rot{}{}\ (\yAcc{A}{B}  - \biasAcc - \noiseAcc{B}) + \gravity{A}, \\
    & \dot{\mathbf{{d}}}_F = \mathbf{Z}_F\ (-\noiseLinVel{F}),  \\
    & \dot{\mathbf{{Z}}}_F = \mathbf{Z}_F\ S(-\noiseAngVel{F}), \\
    & \dot{\mathbf{{\bias}}} = \noiseBias{B}.
    \end{split}
\end{equation}

The overall discrete system dynamics, explicitly written in the form of Eq. \eqref{eq:sec:background:ekf-lie-group-system}, is then described as, \looseness=-1
\begin{align}
\label{eq:sec:estimation:diligent-kio-sys-dynamics}
\begin{split}
\mathbf{p}_{\knext} =&\ \mathbf{p}_{\kcurr} + \Rot{}{}_{\kcurr}( \Rot{}{}_{\kcurr}^T  \mathbf{v}_{\kcurr}  \Delta T  +  \frac{1}{2} \acc{A}{B}_\kcurr \Delta T^2),\\
\Rot{}{}_\knext =&\ \Rot{}{}_\kcurr\ \Exp((\yGyroBar{A}{B}\ -\ \noiseGyro{B} )\ \Delta T), \\
\mathbf{v}_{\knext} =& \; \mathbf{v}_{\kcurr} + \; \Rot{}{}_{\kcurr} \acc{A}{B}_\kcurr  \Delta T ,\\
\mathbf{d}_{F_{\knext}} =&\ \mathbf{d}_{F_{\kcurr}} +\ \mathbf{Z}_{F_{\kcurr}}\ (-\noiseLinVel{F} )\; \Delta T, \\
\mathbf{Z}_{F_{\knext}} =&\ \mathbf{Z}_{F_{\kcurr}} \ \Exp(-\noiseAngVel{F} \; \Delta T), \\
\bias_\knext =&\ \bias_\kcurr + \noiseBias{B}\ \Delta T,
\end{split}
\end{align}
where we have used $\yGyroBar{A}{B}$ and $\accBar{A}{B}$ to denote unbiased gyroscope and accelerometer measurements respectively while  $\acc{A}{B}_\kcurr = \accBar{A}{B}_\kcurr - \noiseAcc{B}$ denotes an unbiased, noise-free accelerometer measurement.

\subsection{Measurement Model}
The measurement model is constructed using the relative pose between the base link and the foot in contact through the forward kinematics map.
The encoder measurements $\encoders = \jointPos + \encoderNoise$ provide the joint positions $\jointPos$ affected by additive white Gaussian noise $\encoderNoise$.
The forward kinematics map $\text{FK}: \R^{\text{DOF}} \mapsto \SE{3}$ provide the measurements  $\Z^F_{\text{SS}} \in \SE{3}$ for single support and $\Z^F_{\text{DS}} \in \SE{3} \times \SE{3}$ for double support. \looseness=-1
\begin{equation}
\begin{split}
\label{eq:sec:estimation:diligent-kio-meas}
& \Z^F_{\text{SS}} =  \text{FK}(\tilde{s})  = \TransformMeasured{B}{F} \in \SE{3} \\
& \Z_{\text{DS}} = \text{blkdiag}(\TransformMeasured{B}{LF},   \TransformMeasured{B}{RF}) \in \SE{3} \times \SE{3}.
\end{split}
\end{equation}

The measurement model can be written in the form of Eq. \eqref{eq:sec:background:ekf-lie-group-meas} as $\TransformMeasured{B}{F} = \Transform{B}{F}\gexphat{\SE{3}}\left(\mathbf{n}_\text{FK}\right)$ where,
\begin{equation}
\label{eq:sec:estimation:diligent-kio-meas-model}
h^F_\text{SS}\left(\X\right) = \begin{bmatrix} {\Rot{}{}^T}{\mathbf{Z}_F} & {\Rot{}{}^T}({\mathbf{d}_F}\ -\ \mathbf{p})   \\
\Zeros{1}{3} & 1
 \end{bmatrix}.
\end{equation}

The left-trivialized forward kinematic noise $\mathbf{n}_\text{FK} \triangleq \text{vec}\left(\fkNoiseLinVel{F}, \fkNoiseAngVel{F}\right) \in \R^6$ is related to the encoder noise $\encoderNoise$ through the left-trivialized relative Jacobian $\relativeJacobianLeftTriv{B}{F}(\jointPos)$.
This can be shown through the push-forward mapping of $\SE{3}$, \looseness=-1
$$
\scalebox{0.85}{
\parbox{.5\linewidth}{$
\Transform{B}{F}\left(\encoders\right) = \Transform{B}{F}\left(\jointPos + \encoderNoise\right) = \Transform{B}{F}\left(\jointPos\right)\ \gexphat{\SE{3}}\left(\relativeJacobianLeftTriv{B}{F}(\jointPos)\ \encoderNoise\right).
$}}
$$
This leads to the measurement noise covariance $\mathbf{N} = \expectation{\mathbf{n}_\text{FK}\ \mathbf{n}_\text{FK}^T} = \relativeJacobianLeftTriv{B}{F}\expectation{\encoderNoise \encoderNoise^T}\relativeJacobianLeftTriv{B}{F}^T$.

\section{Filter Computations}
\label{sec:FILTERS}

In this section, we derive the variants of DILIGENT-KIO based on the choice of the error and time-representation for the system dynamics.
DILIGENT-KIO  is a Discrete Lie Group Extended Kalman Filter for Kinematic Inertial Odometry formulated with the left-invariant error of type 1 and its detailed derivation can be found in \cite{ramadoss2021diligent}.\looseness=-1

\subsection{DILIGENT-KIO-RIE}
DILIGENT-KIO with right-invariant error formulation, DILIGENT-KIO-RIE in short, differs from its original only in terms of the uncertainty management and state reparametrization.
The underlying algorithm is the discrete EKF over matrix Lie groups using the right invariant error of type 1 described in Algorithm \ref{algo:sec:background:dlgekf-rie}.
The discrete system dynamics remain the same as in Eq. \eqref{eq:sec:estimation:diligent-kio-sys-dynamics}, consequently leading to the left trivialized motion model and noise similar to DILIGENT-KIO as in Eqs. \eqref{eq:sec:estimation:diligent-kio-left-triv-motion-model} and \eqref{eq:sec:estimation:diligent-kio-left-triv-motion-noise}, respectively.
It must be noted that the trivialization of the motion model is independent from the choice of error and solely depends on the choice of the system states and their associated dynamics.  

\subsubsection{Left Trivialized Motion Model}
Given the state space representation in Eq. \eqref{eq:sec:estimation:diligent-kio-state}, and  the discrete system dynamics in Eq. \eqref{eq:sec:estimation:diligent-kio-sys-dynamics}, the left trivialized motion model can be obtained as, \looseness=-1
\begin{align}
\begin{split}
\label{eq:sec:estimation:diligent-kio-left-triv-motion-model}
\Omega = \begin{bmatrix}
{\Rot{}{} ^T}{\mathbf{v}} \Delta T +\ \frac{1}{2}\ \accBar{A}{B} \Delta T^2 \\
\yGyroBar{A}{B}\ \Delta T \\ 
\accBar{A}{B}\ \Delta T  \\ 
\Zeros{18}{1}
\end{bmatrix} \in \mathbb{R}^{27},
\end{split}
\end{align}
The discrete, left-trivialized noise vector $\mathbf{w} \in \mathbb{R}^{27}$ becomes,
\begin{align}
\begin{split}
\label{eq:sec:estimation:diligent-kio-left-triv-motion-noise}
 &\text{vec}(-0.5\noiseAcc{B}\Delta T,\ -\noiseGyro{B},\ -\noiseAcc{B},\ \\ &\quad -\noiseLinVel{LF},\ -\noiseAngVel{LF},\  -\noiseLinVel{RF},\ -\noiseAngVel{RF},\ \noiseBias{B} )\Delta T,
\end{split}
\end{align}
with the prediction noise covariance matrix $\mathbf{Q} = \expectation{\mathbf{w} \mathbf{w}^T}$.

\subsubsection{Left Trivialized Motion Model Jacobian}
Given the right-invariant error, the perturbation is induced on the current state estimate from the left side leading to a perturbed state, $\gexphat{\mathcal{M}}({\err_\kprior})\ \Xhat_\kprior$.
The Jacobian of the left trivialized motion model $\mathfrak{F}_\kcurr^R \in \R^{27 \times 27}$ can be computed as, \looseness=-1
\begin{equation}
\label{eq:sec:estimation:diligent-kio-rie-left-triv-motion-jacobian}
\begin{split}
\small
\mathfrak{F}_\kcurr^R &= \frac{\partial}{\partial \err} \Omega\left(\gexphat{\mathcal{M}}({\err_\kprior})\ \Xhat_\kprior, \mathbf{u}_\kcurr\right)_{\err_\kprior = 0} \\
&=
\begingroup
\setlength\arraycolsep{1.6pt}\footnotesize{
\begin{bmatrix}
	\Zero{3} & \frac{1}{2} \Xi_1 \Delta T & \hat{\Rot{}{}}^T \; \Delta T & \Zeros{3}{12} &  -\frac{1}{2}\I{3}\;\Delta T^2 & \Zero{3} \\
	\Zero{3} & \Zero{3} & \Zero{3} & \Zeros{3}{12} &  \Zero{3} & -\I{3}\;\Delta T \\
	\Zero{3} & \Xi_1 & \Zero{3} & \Zeros{3}{12} &  -\I{3}\;\Delta T & \Zero{3} \\
	\Zeros{18}{3} & \Zeros{18}{3} & \Zeros{18}{3} & \Zeros{18}{12} &  \Zeros{18}{3} & \Zeros{18}{3}
	\end{bmatrix}
	}\endgroup,
\end{split}
\end{equation}
where, we have defined $\Xi_1 = \hat{\Rot{}{}}^T S(\gravity{A}) \Delta T$. The linearized error propagation matrix can therefore be computed as $\mathbf{F}_\kcurr = \I{p}\ +\ \gadj{\Xhat_\kprior}\gljac{\mathcal{M}}(\hat{\Omega}_\kcurr) \mathfrak{F}_\kcurr^R$. \looseness=-1

\subsubsection{Measurement Update}
The measurement model follows Eq. \eqref{eq:sec:estimation:diligent-kio-meas-model} and the measurement model Jacobian for single support can be obtained as,
\begin{equation}
\label{eq:sec:estimation:diligent-kio-rie-measmodeljacobianLF}
\begin{split}
\mathbf{H}^{LF} = 
\begingroup
\setlength\arraycolsep{1.8pt}\footnotesize{
\begin{bmatrix} 
-{\mathbf{\hat{Z}}_{LF}^T} & {\mathbf{\hat{Z}}_{LF}^T}S({\mathbf{\hat{d}}_{LF}}) & \Zero{3} & {\mathbf{\hat{Z}}_{LF}^T} & -{\mathbf{\hat{Z}}_{LF}^T}S({\mathbf{\hat{d}}_{LF}}) & \Zeros{3}{12} \\
\Zero{3} & -{\mathbf{\hat{Z}}_{LF}^T}  & \Zero{3} & \Zero{3} & {\mathbf{\hat{Z}}_{LF}^T} & \Zeros{3}{12} 
\end{bmatrix}
}\endgroup , \\
\mathbf{H}^{RF} = 
\begingroup
\setlength\arraycolsep{1.8pt}\footnotesize{
\begin{bmatrix} 
-{\mathbf{\hat{Z}}_{RF}^T} & {\mathbf{\hat{Z}}_{RF}^T}S({\mathbf{\hat{d}}_{RF}}) & \Zeros{3}{9} & {\mathbf{\hat{Z}}_{RF}^T}  & -{\mathbf{\hat{Z}}_{RF}^T}S({\mathbf{\hat{d}}_{RF}}) & \Zeros{3}{6} \\
\Zero{3} & -{\mathbf{\hat{Z}}_{RF}^T}  & \Zeros{3}{9} & \Zero{3} & {\mathbf{\hat{Z}}_{RF}^T} & \Zeros{3}{6} 
\end{bmatrix}
}\endgroup.
\end{split}
\end{equation}

Followed by the correction of the predicted state and the covariance, the update step ends with a state reparametrization $\Xhat_\kest\ =\ \gexphat{G}(\mathbf{m}_\knext^{-})\ \Xhat_\kpred$ and the covariance reparametrization $\cov_\kest\ =\ \gljac{G}(\mathbf{m}_\knext^{-})\ \cov_\kest^{-}\ \gljac{G}(\mathbf{m}_\knext^{-})^T$.

\subsection{CODILIGENT-KIO-RIE}
This is a continuous-discrete EKF on matrix Lie groups with right invariant error formulation and non-invariant observations evolving over matrix Lie groups, in the form of \eqref{eq:sec:background:cdekf-rie}.
In the propagation step, the mean can be simply propagated using $\frac{d}{dt}\Xhat_t = f_{\mathbf{u}_t}\left(\Xhat_t\right) $.
The linearized error dynamics can be used to compute the error propagation matrices which can then used to obtain the predicted state error covariance.
The right-invariant error $\eta^R_\mathcal{M} = \X \Xhat^{-1}$ for elements in state space $\mathcal{M}$ can be written as a tuple,
\begin{equation}
    \label{eq:eq:sec:estimation:cdekf-sys-rie}
    \begin{split}
    \eta^R_\mathcal{M} &=\big(\mathbf{p} - \Rot{}{}\hat{\Rot{}{}}^T \mathbf{\hat{p}}, \
             \Rot{}{}\hat{\Rot{}{}}^T, \ 
             \mathbf{v} - \Rot{}{}\hat{\Rot{}{}}^T \mathbf{\hat{v}}, \ \\
             &  \quad \quad \mathbf{d}_{LF} - \mathbf{Z}_{LF}\mathbf{\hat{Z}}_{LF}^T  \mathbf{\hat{d}}_{LF}, \
             \mathbf{Z}_{LF} \mathbf{\hat{Z}}_{LF}^T, \ \\
             & \quad \quad \mathbf{d}_{RF} - \mathbf{Z}_{RF}\mathbf{\hat{Z}}_{RF}^T  \mathbf{\hat{d}}_{RF}, \
             \mathbf{Z}_{RF} \mathbf{\hat{Z}}_{RF}^T, \
             \bias - \hat{\bias}
             \big)_{\mathcal{M}}.
    \end{split}
\end{equation}
Given, $\eta^R_\mathcal{M} = \gexphat{\mathcal{M}}\left(\err\right) \approx \I{n} + \ghat{\mathcal{M}}{\err}$, the overall non-linear error dynamics $\frac{d}{dt} \eta^R_t = \frac{d}{dt} \left(\X \Xhat^{-1} \right)$ can be computed as, \looseness=-1
\begin{equation}
\label{eq:eq:sec:estimation:cdekf-sys-rie-dynamics}
    \begin{split}
    &  \dot{\eta}_\mathbf{p} =\ \err_\mathbf{v} - S(\mathbf{\hat{p}})\hat{\Rot{}{}}(\err_\biasGyro + \noiseGyro{B}), \\
    &  \dot{\eta}_\mathbf{R} =\ S(\hat{\Rot{}{}} (-\err_\biasGyro - \noiseGyro{B}) ) ,\\
    & \dot{\eta}_\mathbf{v} =\ S(\gravity{A}) \err_\mathbf{R} - S(\mathbf{\hat{v}})\hat{\Rot{}{}}(\err_\biasGyro + \noiseGyro{B}) -\hat{\Rot{}{}}(\err_\biasAcc + \noiseAcc{B}), \\
    & \dot{\eta}_{\mathbf{d}_F} =\ -S(\mathbf{\hat{d}}_F)\hat{\mathbf{Z}}_F\noiseAngVel{F} - \hat{\mathbf{Z}}_F \noiseLinVel{F},\\
    & \dot{\eta}_{\mathbf{Z}_F} =\ S(-\hat{\mathbf{Z}}_F\noiseAngVel{F}),\\
    & \dot{\eta}_\bias =\ \noiseBias{B}.
    \end{split}
\end{equation}

Owing to the log-linearity property, the linearized error dynamics then become $\dot{\err} = \mathbf{F}_c \err - \mathbf{L}_c \mathbf{w}$,  where,
\begin{equation}
\mathbf{F}_c =
\begingroup
\setlength\arraycolsep{1.6pt}\footnotesize{
\begin{bmatrix}
    \Zero{3} & \Zero{3} & \I{3} & \Zeros{3}{12} & \Zero{3} & - S(\mathbf{\hat{p}})\hat{\Rot{}{}} \\
    \Zero{3} & \Zero{3} & \Zero{3} &  \Zeros{3}{12} & \Zero{3} & -\hat{\Rot{}{}} \\
    \Zero{3} & S(\gravity{A}) & \Zero{3} &  \Zeros{3}{12} & -\hat{\Rot{}{}} & - S(\mathbf{\hat{v}})\hat{\Rot{}{}} \\
    \Zeros{18}{3} & \Zeros{18}{3} & \Zeros{18}{3} &  \Zeros{18}{12} & \Zeros{18}{3} &  \Zeros{18}{3}
\end{bmatrix},
}\endgroup
\label{eq:codiligent-kio-rie-error-prop-matrix}
\end{equation}

with the noise vector defined as, 
\begin{equation}
\label{eq:sec:estimation:cd-ekf-rie-prop-noise}
\mathbf{w} \triangleq \text{vec}(\Zeros{3}{1},\ \noiseGyro{B}, \noiseAcc{B}, \noiseLinVel{LF}, \noiseAngVel{LF}, \noiseLinVel{RF}, \noiseAngVel{RF}, -\noiseBias{B} ).
\end{equation}
The matrix $\mathbf{L}_c$ is given by the adjoint matrix $\gadj{\Xhat}$ of the state representation $\mathcal{M}$.
When the biases are not considered within the system, the error propagation matrix $\mathbf{F}_c$ becomes time-invariant (first 21 rows and columns of $\mathbf{F}_c$), since the system defined in Eq. \eqref{eq:sec:background:cdekf-rie} obeys the property of group-affine dynamics in Eq. \eqref{eq:sec:background:invekf-group-affine-dynamics} leading to an autonomous error propagation.
Nevertheless, when the biases are considered, the overall error system is only dependent on the state trajectory through the noise and bias errors.
The continuous time filter equations for the propagation step becomes, \looseness=-1
\begin{equation}
    \begin{split}
        & \frac{d}{dt} \Xhat_t = f_{\mathbf{u}_t}(\Xhat_t), \\
        &  \frac{d}{dt} \cov_t = \mathbf{F}_c \cov_t + \cov_t \mathbf{F}_c^T + \mathbf{\hat{Q}}_c.
    \end{split}
\end{equation}
where $\mathbf{\hat{Q}}_c = \mathbf{L}_c \text{Cov}(\mathbf{w}) \mathbf{L}_c^T$. For the software implementation of the continuous-time equations, we discretize the continuous dynamics using a zero-order hold on the inputs with a sampling period $\Delta T$. The discrete system dynamics takes the form of Eq. \eqref{eq:sec:estimation:diligent-kio-sys-dynamics}, while the uncertainty propagation is approximated as, \looseness=-1
\begin{equation}
    \begin{split}
        & \mathbf{F}_k = \exp(\mathbf{F}_c \Delta T) \approx \I{p} + \mathbf{F}_c \Delta T, \\
        & \mathbf{Q}_k = \mathbf{F}_k \mathbf{\hat{Q}}_c \mathbf{F}_k^T \Delta T,\\
        & \mathbf{P}_\kpred = \F_\kcurr\ \cov_\kprior\ \F_\kcurr^T + \Q_\kcurr.
    \end{split}
\end{equation}

The filter observations are the same as those defined in Eq. \eqref{eq:sec:estimation:diligent-kio-meas-model}, however due to the choice of right invariant error, the measurement update and the state reparametrization follows the procedure defined for DILIGENT-KIO-RIE, with the measurement model $\mathbf{H}_\knext$ defined in Eq. \eqref{eq:sec:estimation:diligent-kio-rie-measmodeljacobianLF}.

\subsection{CODILIGENT-KIO}
CODILIGENT-KIO is a left invariant error counterpart of CODILIGENT-KIO-RIE.
The left-invariant error $\eta^L_\mathcal{M} = \Xhat^{-1} \X $ can be written as,
\begin{equation}
    \label{eq:eq:sec:estimation:cdekf-sys-lie}
    \begin{split}
    \eta^L_\mathcal{M} &=\big( \hat{\Rot{}{}}^T(\mathbf{p} - \mathbf{\hat{p}}), \
             \hat{\Rot{}{}}^T \Rot{}{}, \ 
             \hat{\Rot{}{}}^T(\mathbf{v} - \mathbf{\hat{v}}), \ \\
             &  \quad \quad \mathbf{\hat{Z}}_{LF}^T(\mathbf{d}_{LF} - \mathbf{\hat{d}}_{LF}), \
             \mathbf{\hat{Z}}_{LF}^T \mathbf{Z}_{LF}, \ \\
             & \quad \quad \mathbf{\hat{Z}}_{RF}^T (\mathbf{d}_{RF} - \mathbf{\hat{d}}_{RF}), \
             \mathbf{\hat{Z}}_{RF}^T \mathbf{Z}_{RF} , \
             \bias - \hat{\bias}
             \big)_{\mathcal{M}}.
    \end{split}
\end{equation}
Given, $\eta^L_\mathcal{M} = \gexphat{\mathcal{M}}\left(\err\right) \approx \I{n} + \ghat{\mathcal{M}}{\err}$, the overall non-linear error dynamics $\frac{d}{dt} \eta^L_t = \frac{d}{dt} \left(\Xhat^{-1} \X \right)$ can be written in terms of the perturbation vector $\err$ as,
\begin{equation}
\label{eq:eq:sec:estimation:cdekf-sys-lie-dynamics}
    \begin{split}
    &  \dot{\eta}_\mathbf{p} =\ \err_\mathbf{v} - S(\yGyro{A}{B} - \mathbf{\hat{b}}_g)\err_\mathbf{p}, \\
    &  \dot{\eta}_\mathbf{R} =\ S(-\err_\biasGyro - \noiseGyro{B} ), \\
    & \dot{\eta}_\mathbf{v} =\   - S({\yGyro{A}{B}} - \biasGyroHat) \err_\mathbf{v} - S(\yAcc{A}{B} - \biasAccHat)  \err_\mathbf{R} -\err_\biasAcc -\noiseAcc{B}, \\
    & \dot{\eta}_{\mathbf{d}_F} =\ - \noiseLinVel{F},\\
    & \dot{\eta}_{\mathbf{Z}_F} =\ -S(\noiseAngVel{F}),\\
    & \dot{\eta}_\bias =\ \noiseBias{B}.
    \end{split}
\end{equation}

Given the noise vector in Eq. \eqref{eq:sec:estimation:cd-ekf-rie-prop-noise}, the linearized error dynamics then can be written as $\dot{\err} = \mathbf{F}_c \err - \mathbf{L}_c \mathbf{w}$ with $\mathbf{L}_c = \I{27}$. 
Considering $\yGyroBar{A}{B} = \yGyro{A}{B} - \biasGyroHat$ and $\accBar{A}{B} = \yAcc{A}{B} - \biasAccHat$, the linearized error propagation matrix $\mathbf{F}_c$ becomes, \looseness=-1
\begin{equation}
\mathbf{F}_c =
\begingroup
\setlength\arraycolsep{1pt}\footnotesize{
\begin{bmatrix}
    -S(\yGyroBar{A}{B}) & \Zero{3} & \I{3} & \Zeros{3}{12} & \Zero{3} & \Zero{3} \\
    \Zero{3} & \Zero{3} & \Zero{3} &  \Zeros{3}{12} & \Zero{3} & -\I{3} \\
    \Zero{3} & -S(\accBar{A}{B}) & -S(\yGyroBar{A}{B}) &  \Zeros{3}{12} & -\I{3} & \Zero{3} \\
    \Zeros{18}{3} & \Zeros{18}{3} & \Zeros{18}{3} &  \Zeros{18}{12} & \Zeros{18}{3} &  \Zeros{18}{3}
\end{bmatrix}
}\endgroup
\end{equation}

It can be observed that, for CODILIGENT-KIO, the error propagation depends not only on the bias errors and the noise terms, but also the IMU measurements compensated by estimated IMU biases, leading to a time-varying linearized error system. \looseness=-1
The measurement model Jacobians during single support is the same as the one obtained for DILIGENT-KIO, \looseness=-1
\begin{equation}
\label{eq:sec:estimation:diligent-kio-measmodeljacobianLF}
\begin{split}
\mathbf{H}^{LF} = 
\begingroup
\setlength\arraycolsep{1.8pt}\footnotesize{
\begin{bmatrix} 
-{\mathbf{\hat{Z}}_{LF}^T \hat{\Rot{}{}}} & -{\mathbf{\hat{Z}}_{LF}^T}S({\mathbf{\hat{p}} - \mathbf{\hat{d}}_{LF}}){\hat{\Rot{}{}}} & \Zero{3} & \I{3} & \Zero{3} & \Zeros{3}{12} \\
\Zero{3} & -{\mathbf{\hat{Z}}_{LF}^T \hat{\Rot{}{}}}  & \Zero{3} & \Zero{3} & \I{3} & \Zeros{3}{12} 
\end{bmatrix}}\endgroup, \\
\mathbf{H}^{RF} = \begingroup
\setlength\arraycolsep{1.8pt}\footnotesize{\begin{bmatrix} 
-{\mathbf{\hat{Z}}_{RF}^T \hat{\Rot{}{}}} & -{\mathbf{\hat{Z}}_{RF}^T}S({\mathbf{\hat{p}} - \mathbf{\hat{d}}_{RF}}){\hat{\Rot{}{}}} & \Zeros{3}{9} & \I{3} & \Zero{3} & \Zeros{3}{6} \\
\Zero{3} & -{\mathbf{\hat{Z}}_{RF}^T \hat{\Rot{}{}}}  & \Zeros{3}{9} & \Zero{3} & \I{3} & \Zeros{3}{6} 
\end{bmatrix}}\endgroup.
\end{split}
\end{equation}

These matrices are stacked together during the double support.
The filter observations and the state reprametrization equations are the same as those defined for DILIGENT-KIO. \looseness=-1
\section{Experimental Results}
\label{sec:RESULTS}

In this section, we present the results from the comparison of flat-foot filters listed in Table \ref{table:soa} for experiments conducted on the iCub humanoid platform. \looseness=-1

\subsection{Experimental Setup}
We conduct real-world experiments on iCub v2.5 platform while simulated experiments are carried out on iCub v3.0 platform.
For both these robots, a subset of 26 degrees-of-freedom equipped with joint encoders is used to measure the joint angles at $1000 \si{Hz}$. iCub v2.5 is equipped with an XSens MTi-300 series IMU mounted in its base link streaming measurements at $100 \si{Hz}$, while a Gazebo simulated IMU mounted on the chest link is used for iCub v3.0.
The contact states are inferred through a Schmitt trigger thresholding of contact wrenches estimated by the estimation algorithm presented in \cite{nori2015icubWBC} available at $100 \si{Hz}$.
The filters are run at $100 \si{Hz}$, and the encoder measurements are sub-sampled to the same frequency as the estimator. The ground truth pose measurements on the real robot are obtained from the Vicon motion capture system. \looseness=-1

\subsection{Flat-foot filters for comparison}
We compare the state-of-the-art filters, OCEKF \cite{rotella2014state}, InvEKF-F \cite{qin2020novel}, and DILIGENT-KIO \cite{ramadoss2021diligent} along with those proposed in this article, DILIGENT-KIO-RIE, CODILIGENT-KIO and CODILIGENT-KIO-RIE.
OCEKF imposes constraints on the rank of the null space of the observability matrix to choose linearization points for the filter computations in order to maintain consistency.
InvEKF-F exploits the Lie group geometry to provide an implicit consistency for the filter owing to the autonomous error evolution property and invariant observation structure. 
DILIGENT-KIO and DILIGENT-KIO-RIE neither have an autonomous error evolution nor use invariant observations leading to a lack of inherent consistency properties.
CODILIGENT-KIO and CODILIGENT-KIO-RIE follow an autonomous error propagation when biases are not considered, while retaining non-invariant observations evolving over matrix Lie groups. 
We use the same noise parameters and initial state standard deviations as shown in Table \ref{table:parameters} for all the filters across all experiments. \looseness=-1

\begin{table}
	\caption{Noise parameters and prior deviations}
	\label{table:parameters}
	\begin{center}
		\tabcolsep=0.02cm
		\scalebox{0.8}{
		\begin{tabular}{cc}	
						\begin{tabular}{c|c}
						\hline 
							Sensor & noise std dev. \\
								\hline	
							Lin. Accelerometer & 0.09 \si{\metre \per {\second^2}}	 \\
							Gyroscope & 0.01 \si{\radian \per \second} \\
							Acc. bias & 0.01 \si{\metre \per {\second^2}} \\
							Gyro. bias &  0.001 \si{\radian \per \second} \\
							Contact foot lin. velocity & 0.009 \si{\meter\per\second}\\
							Contact foot ang. velocity & 0.004 \si{\radian \per \second} \\							
							Joint encoders & 0.1 \si{\degree}						\\	\hline 
						\end{tabular}
&
					\begin{tabular}{c|c}
					\hline 
						State element & initial std dev. \\
						\hline	
						IMU \& feet position & 0.01 \si{\meter} \\						
						IMU \& feet orientation & 10 \si{\degree} \\
						IMU linear velocity & 0.5 \si{\meter\per\second} \\
						Acc. bias & 0.01 \si{\metre \per {\second^2}} \\
						Gyro. bias & 0.002  \si{\radian \per \second}
						\\	\hline 
					\end{tabular}						
		\end{tabular}
		}
	\end{center}
	\vspace{-6mm}
\end{table}

\subsection{Discussion}
\begin{figure*}
	\begin{subfigure}{0.49\textwidth}
	\centering
\includegraphics[scale=0.24]{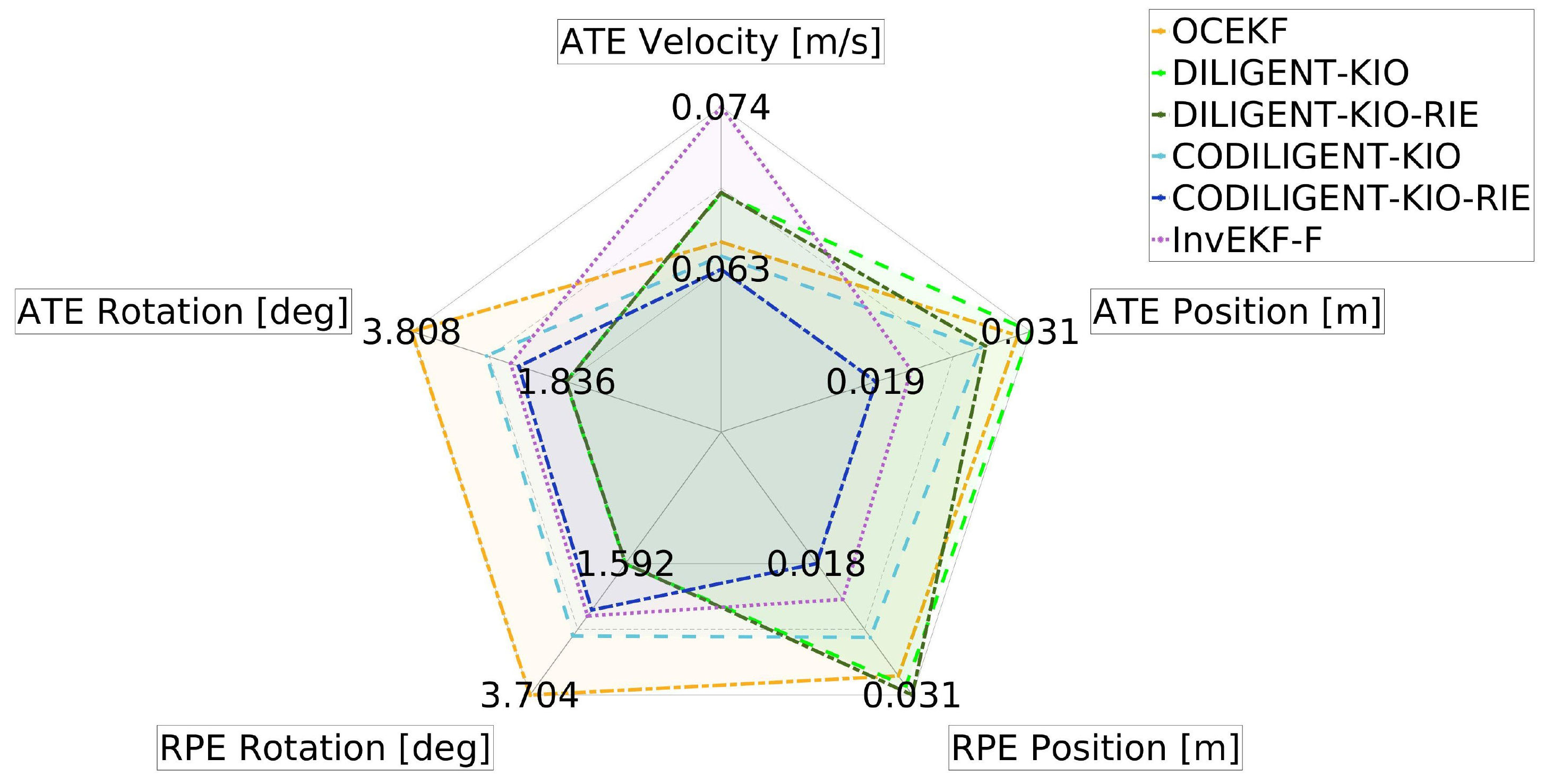}
	\end{subfigure}
\begin{subfigure}{0.49\textwidth}
	\centering
\includegraphics[scale=0.24]{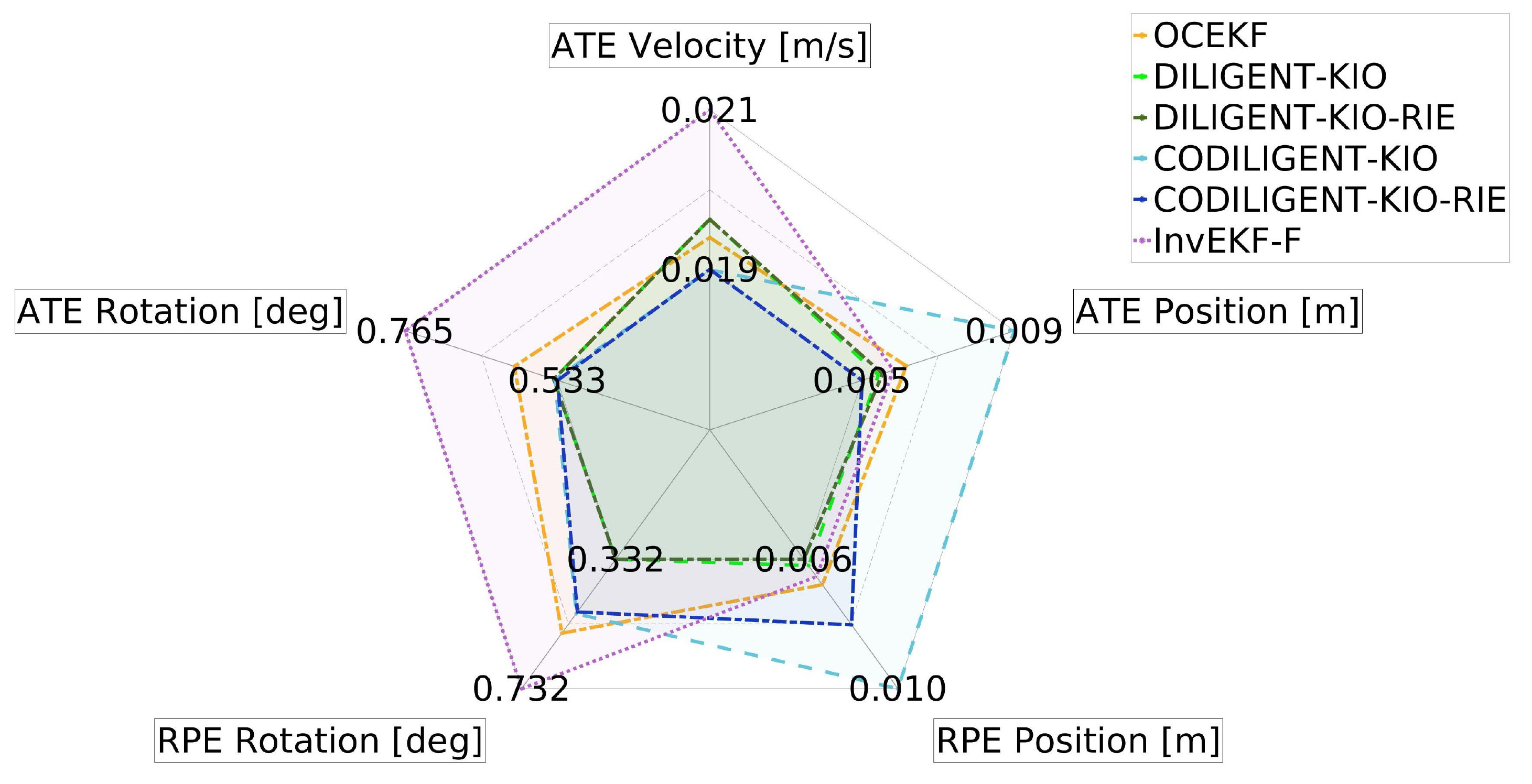}
\end{subfigure}	
    \caption{Absolute Trajectory Error and Relative Pose Error of the base link trajectory for a real-world, $1 \si{\meter}$ forward walking experiment (left) and COM sinusoid trajectory tracking experiment (right) on iCub v2.5 computed for OCEKF (yellow), DILIGENT-KIO (light green), DILIGENT-KIO-RIE (dark green), CODILIGENT-KIO (light blue), CODILIGENT-KIO-RIE (dark blue), and InvEKF-F (purple). \looseness=-1}
    \label{fig:errors-walking1m-comsinusoid-icubV2_5}
\end{figure*}

\begin{figure}
		\centering
\includegraphics[scale=0.24]{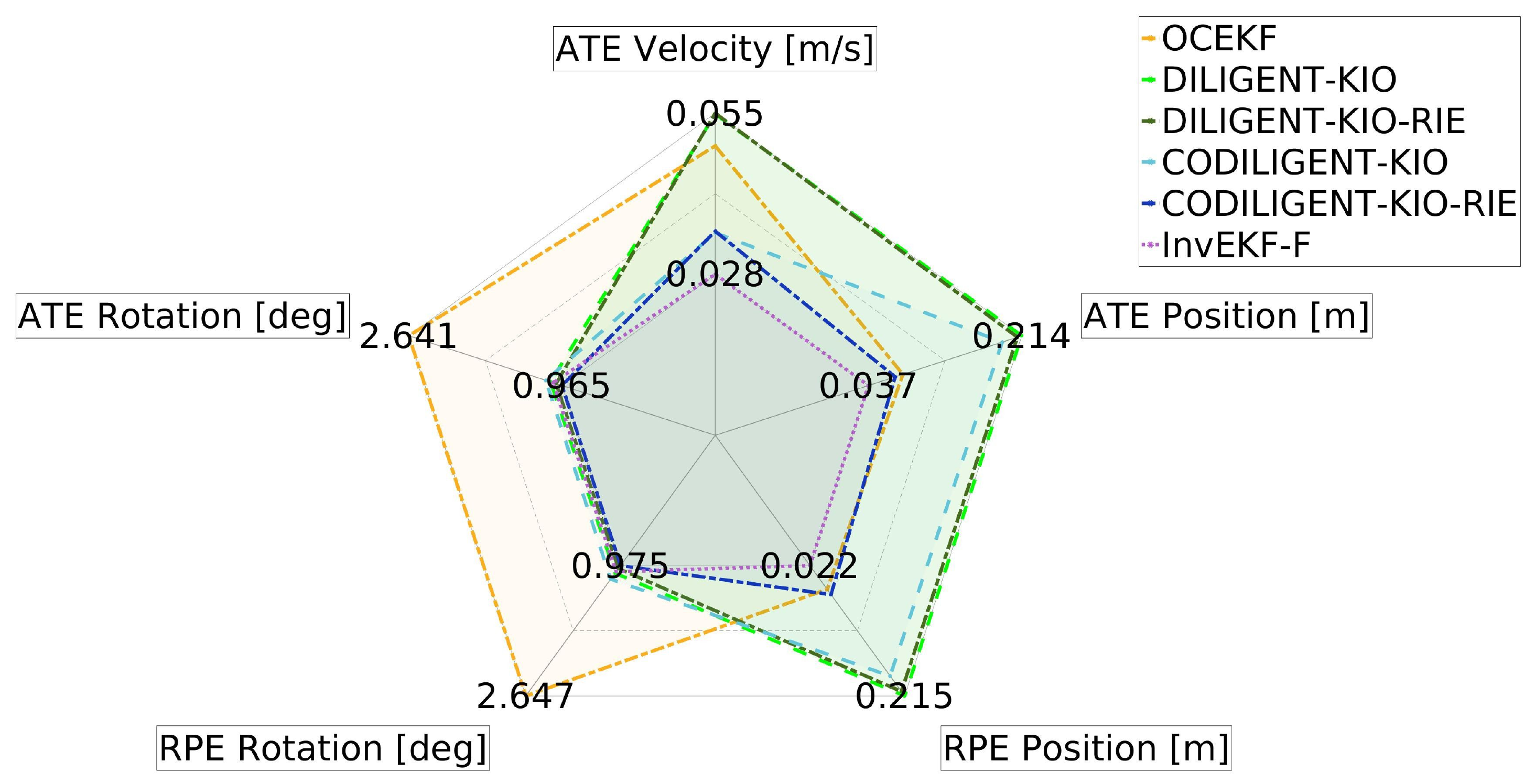}
    \caption{Absolute Trajectory Error and Relative Pose Error the base link trajectory for a Gazebo-simulated, $3 \si{\meter}$ forward walking experiment on iCub v3.0. \looseness=-1}
    \label{fig:errors-walking3m-icubGazeboV3}
\end{figure}

We rely on Absolute Trajectory Error (ATE) and Relative Pose Error (RPE) \cite{sturm2012benchmark} computed from Vicon and estimated trajectories for the evaluation.
Figures \ref{fig:errors-walking1m-comsinusoid-icubV2_5} show the base state error comparison for a position-controlled, $1 \si{\meter}$ forward walking  [20]  experiment  and  a torque-controlled Center of Mass (CoM) sinusoidal trajectory tracking experiment [16] conducted on the  real  iCub v2.5 robotic  platform. 
Figure \ref{fig:errors-walking3m-icubGazeboV3} depicts the comparison for a $3 \si{\meter}$ forward walking experiment on a Gazebo simulated iCub v3 platform. \looseness=-1

It can be observed that the filters on matrix Lie groups generally suffer from lesser orientation errors.
For the shorter-duration walking and COM sinusoid experiment conducted on the real robot, the variants of DILIGENT-KIO suffer from least errors along the observable directions of orientation and velocity.
These filters, with the exception of CODILIGENT-KIO-RIE, are seen to be affected by an increase of position error from $3$cm for $1 \si{\meter}$ walking to a $20$cm for $3 \si{\meter}$ walking.
While OCEKF is seen to suffer less from position errors, InvEKF-F and CODILIGENT-KIO-RIE show an overall consistent performance across all the experiments.
\vspace{2em}
\begin{figure*}[t!]
\begin{subfigure}{0.33\textwidth}
		\centering
\includegraphics[scale=0.2]{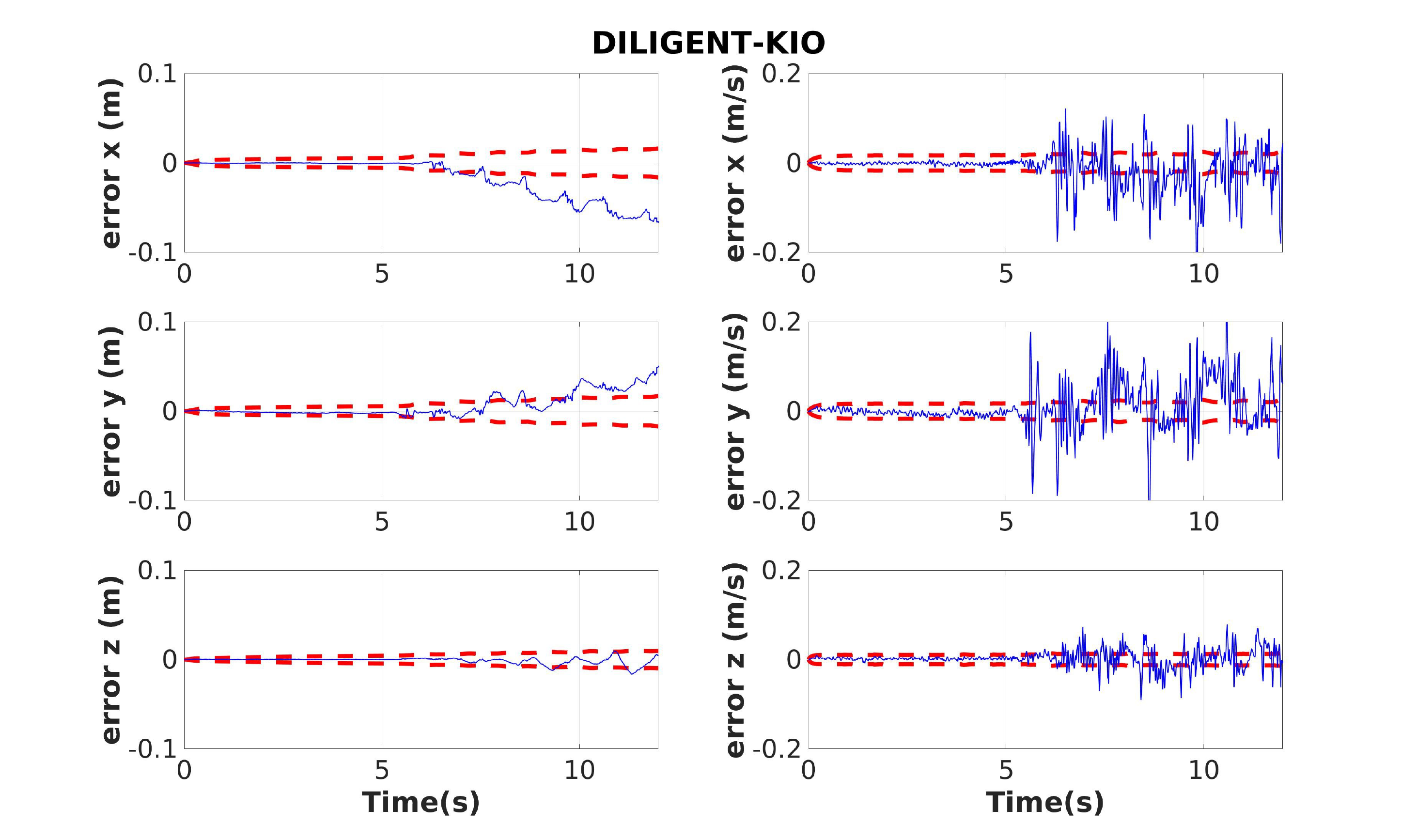}
\end{subfigure}
\begin{subfigure}{0.33\textwidth}
		\centering
\includegraphics[scale=0.2]{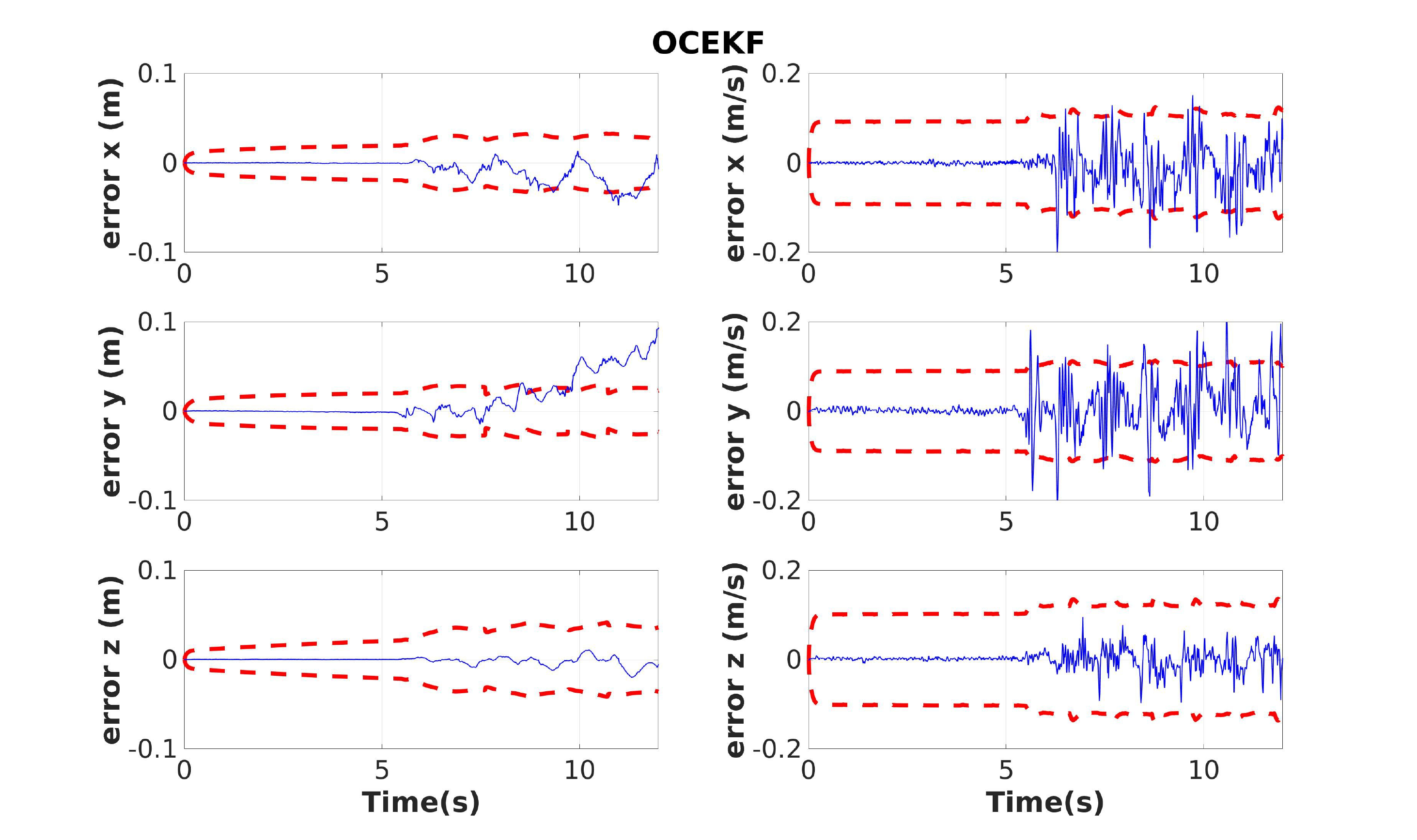}
	\end{subfigure}
\begin{subfigure}{0.33\textwidth}
		\centering
\includegraphics[scale=0.2]{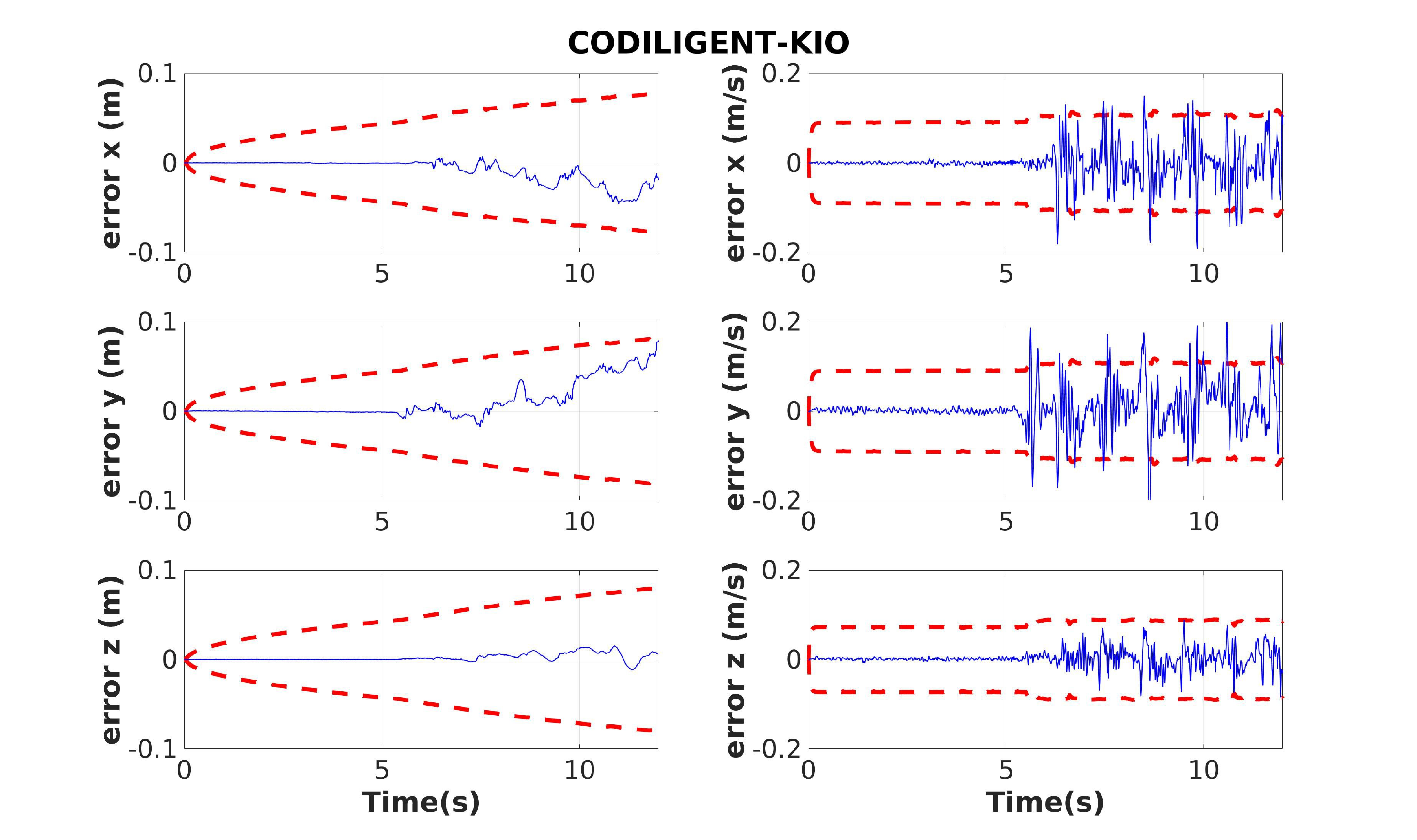}
	\end{subfigure}
\begin{subfigure}{0.33\textwidth}
		\centering
\includegraphics[scale=0.2]{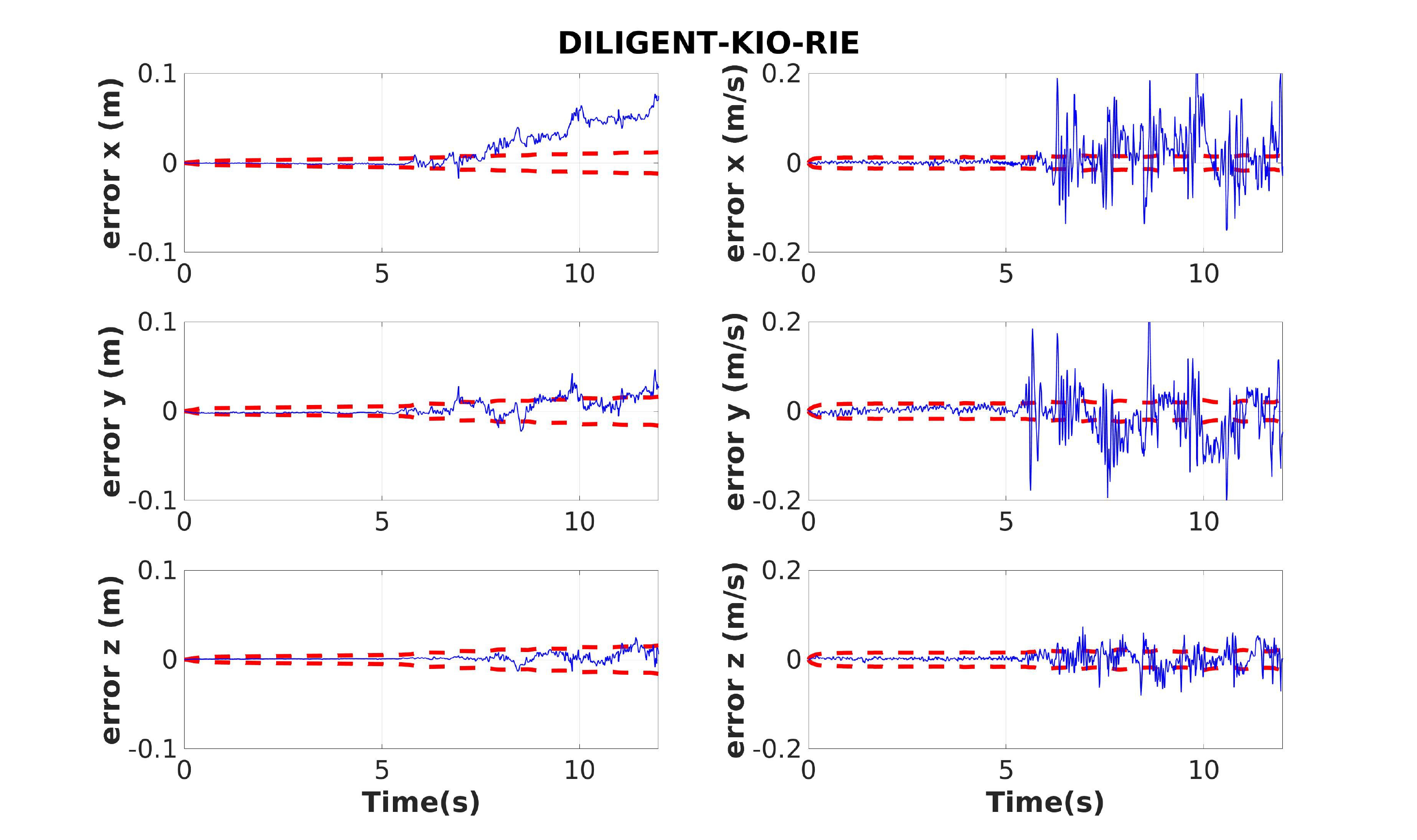}
	\end{subfigure}
\begin{subfigure}{0.33\textwidth}
		\centering
\includegraphics[scale=0.2]{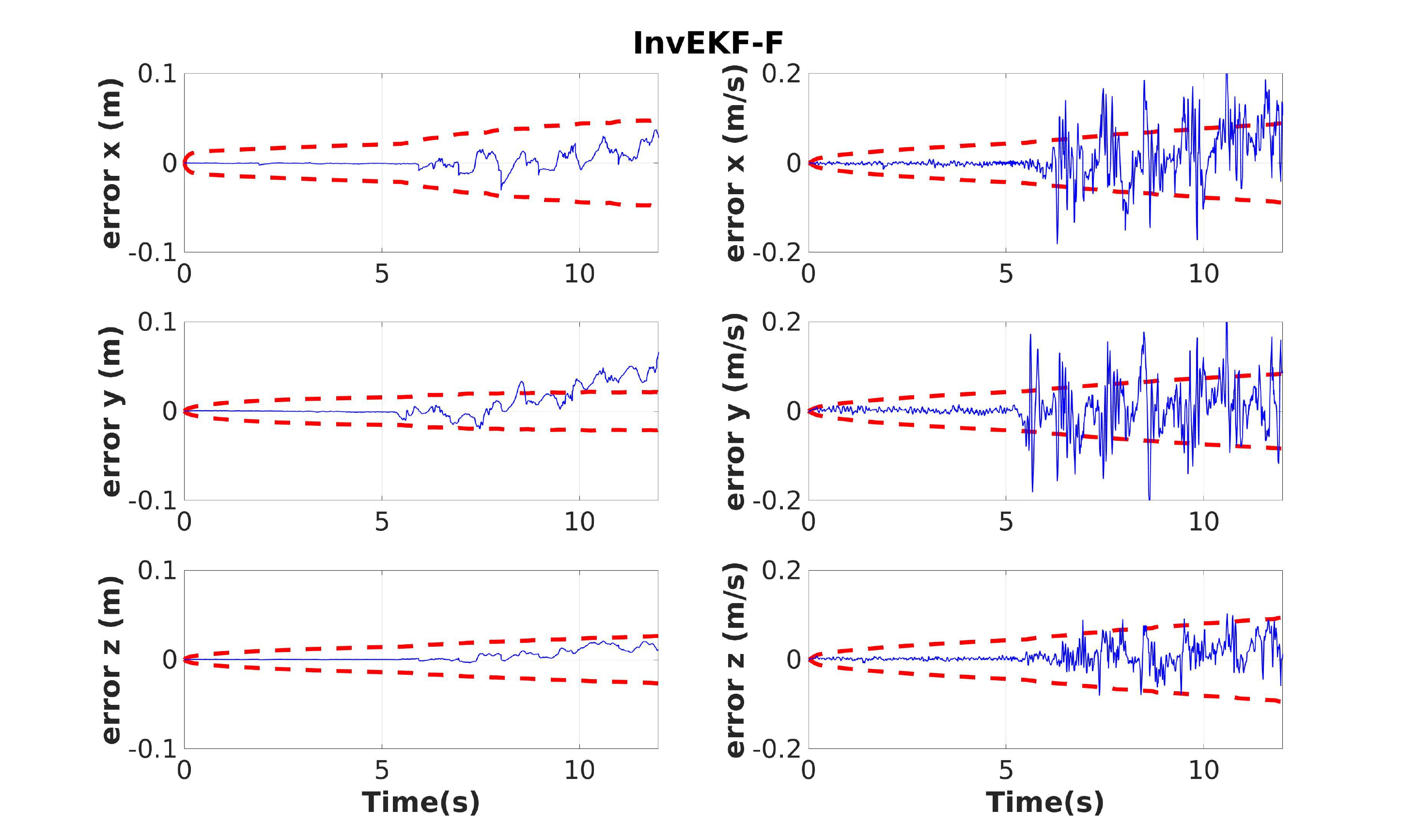}
	\end{subfigure}
\begin{subfigure}{0.33\textwidth}
		\centering
\includegraphics[scale=0.2]{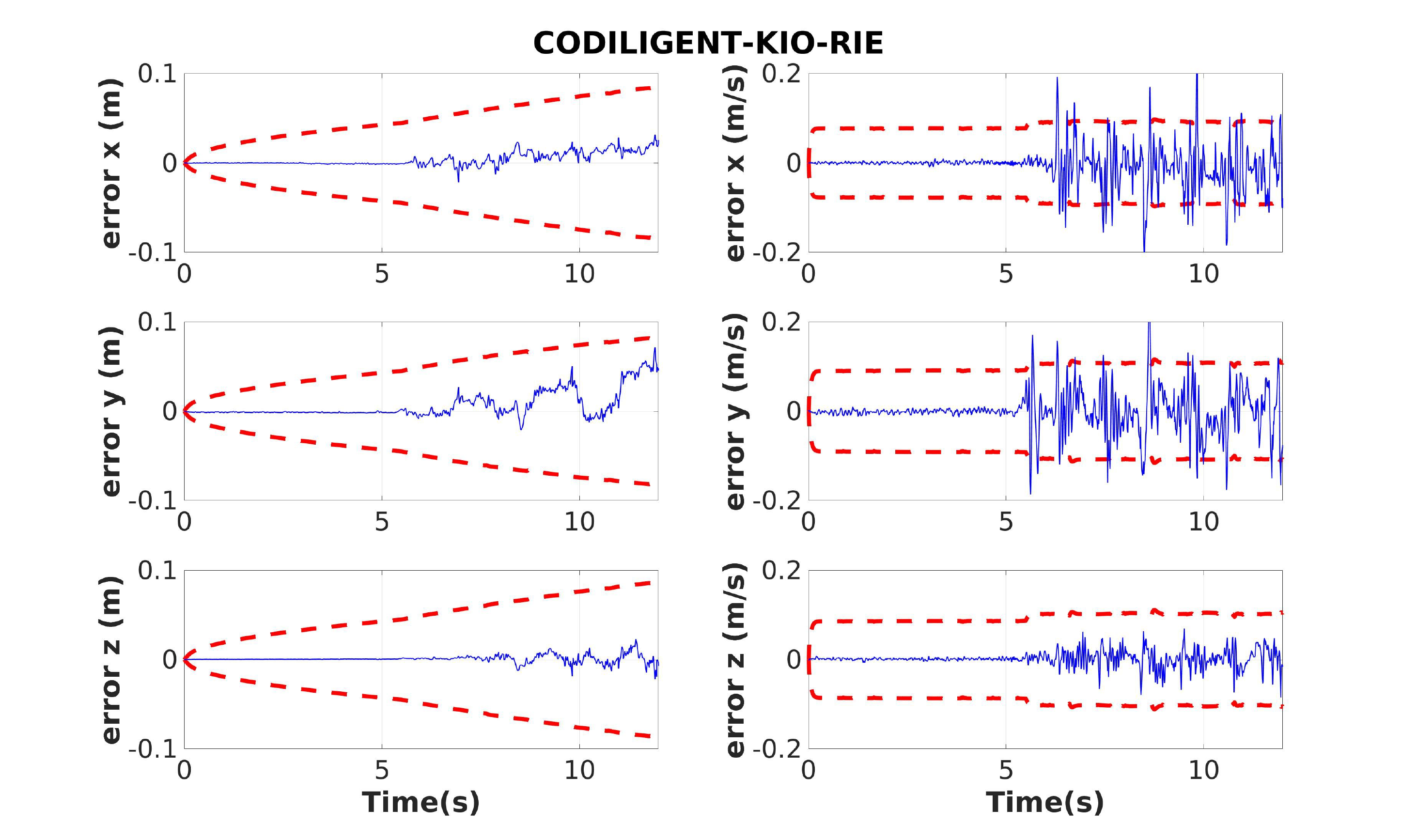}
\end{subfigure}	
\caption{Comparison of evolution of position and velocity errors (blue lines) respectively along with 99$\%$ estimated uncertainty envelope (red lines) of flat-foot filters during a 1m forward-walking experiment conducted on a real iCub humanoid platform.}
\label{fig:chap:diligent-variants-comparison}
\end{figure*}

The plots in Figure \ref{fig:chap:diligent-variants-comparison} show the error (in blue) in the unobservable position directions and observable velocity directions for a 1m forward walking experiment conducted on the real robot.
They also show a 99$\%$ estimated uncertainty envelope (in red) obtained from the covariance computed by the filters.
The uncertainty envelope is bound to increase over time in the unobservable directions, while it converges to track the error in the observable directions.
It can be seen that for DILIGENT-KIO and DILIGENT-KIO-RIE, the errors in both the unobservable position direction and observable velocity directions do not remain within the 99$\%$ estimated uncertainty envelope of the filter.
This shows that both these filters become overconfident in their estimation.
InvEKF-F and OCEKF are seen to cope considerably well leading to the errors being bounded within the envelope.
CODILIGENT-KIO and CODILIGENT-KIO-RIE perform better than their discrete counterparts leading to a consistent tracking of error statistics. \looseness=-1

The overall better performance of InvEKF-F and CODILIGENT-KIO-RIE is related to the time-invariant error-propagation matrix (Eq. \eqref{eq:codiligent-kio-rie-error-prop-matrix}), when the biases are negligible, while the former retains also a time-invariant measurement Jacobian.
These filters obtain these properties with the choice of right-invariant error and group-affine system dynamics.
However, to improve the reliability of all the variants of DILIGENT-KIO, a thorough observability and consistency analysis is required, since their linearized error system are dependent on the state trajectory or the IMU measurements.

\section{Conclusion}
\label{sec:CONCLUSION}
In this paper, we compare EKF-based flat foot filters for floating base estimation of humanoid robots, based on the choice of state and measurement representation, error-formulation and system dynamics representation.
In order to aid the comparison with the state-of-the-art \emph{consistent} filters such as observability-rules based OCEKF \cite{rotella2014state} and invariant filtering based InvEKF-F \cite{qin2020novel}, we derive a few variants of previously proposed DILIGENT-KIO \cite{ramadoss2021diligent} based on the aforementioned modeling choices for estimator design.
We notice that, the filters designed on matrix Lie groups that exploit a group-affine system dynamics structure tend to perform better as consistent estimators than the standard EKF and OCEKF counterparts. 
Future works may exploit the group-affine property for discrete-time systems \cite{barrau2018invariant} to extend DILIGENT-KIO to achieve autonomous error propagation. \looseness=-1

\addtolength{\textheight}{-0.215cm}   
\bibliography{bibliography}

\begin{thebibliography}{10}
\providecommand{\url}[1]{#1}
\csname url@rmstyle\endcsname
\providecommand{\newblock}{\relax}
\providecommand{\bibinfo}[2]{#2}
\providecommand\BIBentrySTDinterwordspacing{\spaceskip=0pt\relax}
\providecommand\BIBentryALTinterwordstretchfactor{4}
\providecommand\BIBentryALTinterwordspacing{\spaceskip=\fontdimen2\font plus
\BIBentryALTinterwordstretchfactor\fontdimen3\font minus
  \fontdimen4\font\relax}
\providecommand\BIBforeignlanguage[2]{{%
\expandafter\ifx\csname l@#1\endcsname\relax
\typeout{** WARNING: IEEEtran.bst: No hyphenation pattern has been}%
\typeout{** loaded for the language `#1'. Using the pattern for}%
\typeout{** the default language instead.}%
\else
\language=\csname l@#1\endcsname
\fi
#2}}

\bibitem{bloesch2013state}
M.~Bloesch, M.~Hutter, M.~A. Hoepflinger, S.~Leutenegger, C.~Gehring, C.~D.
  Remy, and R.~Siegwart, ``State estimation for legged robotsconsistent fusion
  of leg kinematics and imu,'' \emph{Robotics}, vol.~17, pp. 17--24, 2013.

\bibitem{rotella2014state}
N.~Rotella, M.~Bloesch, L.~Righetti, and S.~Schaal, ``State estimation for a
  humanoid robot,'' in \emph{2014 {IEEE}/{RSJ} International Conference on
  Intelligent Robots and Systems}.\hskip 1em plus 0.5em minus 0.4em\relax
  {IEEE}, sep 2014.

\bibitem{hartley2020contact}
R.~Hartley, M.~Ghaffari, R.~M. Eustice, and J.~W. Grizzle, ``Contact-aided
  invariant extended kalman filtering for robot state estimation,'' \emph{The
  International Journal of Robotics Research}, vol.~39, no.~4, pp. 402--430,
  2020.

\bibitem{barrau2017invariant}
A.~Barrau and S.~Bonnabel, ``The invariant extended kalman filter as a stable
  observer,'' vol.~62, no.~4, pp. 1797--1812, apr 2017.

\bibitem{qin2020novel}
M.~Qin, Z.~Yu, X.~Chen, Q.~Huang, C.~Fu, A.~Ming, and C.~Tao, ``A novel foot
  contact probability estimator for biped robot state estimation,'' in
  \emph{2020 IEEE International Conference on Mechatronics and Automation
  (ICMA)}.\hskip 1em plus 0.5em minus 0.4em\relax IEEE, 2020, pp. 1901--1906.

\bibitem{ramadoss2021diligent}
P.~Ramadoss, G.~Romualdi, S.~Dafarra, F.~J. Andrade~Chavez, S.~Traversaro, and
  D.~Pucci, ``Diligent-kio: A proprioceptive base estimator for humanoid robots
  using extended kalman filtering on matrix lie groups,'' in \emph{2021 IEEE
  International Conference on Robotics and Automation (ICRA)}, 2021, pp.
  2904--2910.

\bibitem{huang2010observability}
G.~P. Huang, A.~I. Mourikis, and S.~I. Roumeliotis, ``Observability-based rules
  for designing consistent ekf slam estimators,'' \emph{The International
  Journal of Robotics Research}, vol.~29, no.~5, pp. 502--528, 2010.

\bibitem{traversaro2019multibody}
S.~Traversaro and A.~Saccon, ``Multibody dynamics notation (version 2),'' 2019.

\bibitem{chirikjian2011stochastic}
G.~S. Chirikjian, \emph{Stochastic Models, Information Theory, and Lie Groups,
  Volume 2: Analytic Methods and Modern Applications}.\hskip 1em plus 0.5em
  minus 0.4em\relax Springer Science \& Business Media, 2011, vol.~2.

\bibitem{bourmaud2015continuous}
G.~Bourmaud, R.~M{\'e}gret, M.~Arnaudon, and A.~Giremus, ``Continuous-discrete
  extended kalman filter on matrix lie groups using concentrated gaussian
  distributions,'' \emph{Journal of Mathematical Imaging and Vision}, vol.~51,
  no.~1, pp. 209--228, 2015.

\bibitem{barfoot2014associating}
T.~D. Barfoot and P.~T. Furgale, ``Associating uncertainty with
  three-dimensional poses for use in estimation problems,'' \emph{IEEE
  Transactions on Robotics}, vol.~30, no.~3, pp. 679--693, 2014.

\bibitem{nori2015icubWBC}
\BIBentryALTinterwordspacing
F.~Nori, S.~Traversaro, J.~Eljaik, F.~Romano, A.~Del~Prete, and D.~Pucci,
  ``icub whole-body control through force regulation on rigid non-coplanar
  contacts,'' \emph{Frontiers in Robotics and AI}, vol.~2, p.~6, 2015.
  [Online]. Available:
  \url{https://www.frontiersin.org/article/10.3389/frobt.2015.00006}
\BIBentrySTDinterwordspacing

\bibitem{sturm2012benchmark}
J.~Sturm, N.~Engelhard, F.~Endres, W.~Burgard, and D.~Cremers, ``A benchmark
  for the evaluation of rgb-d slam systems,'' in \emph{2012 IEEE/RSJ
  International Conference on Intelligent Robots and Systems}.\hskip 1em plus
  0.5em minus 0.4em\relax IEEE, 2012, pp. 573--580.

\bibitem{barrau2018invariant}
A.~Barrau and S.~Bonnabel, ``Invariant kalman filtering,'' \emph{Annual Review
  of Control, Robotics, and Autonomous Systems}, vol.~1, pp. 237--257, 2018.

\end{thebibliography}
\bibliographystyle{IEEEtran}


\end{document}